\documentclass[11pt]{article}

\usepackage[margin=1in]{geometry}
\usepackage{graphicx}%
\usepackage{multirow}%
\usepackage{amsmath,amssymb,amsfonts}%
\usepackage{amsthm}%
\usepackage{mathrsfs}%
\usepackage[title]{appendix}%
\usepackage{xcolor}%
\usepackage{textcomp}%
\usepackage{manyfoot}%
\usepackage{booktabs}%
\usepackage{algorithm}%
\usepackage{csquotes}
\usepackage{algorithmicx}%
\usepackage{algpseudocode}%
\usepackage{listings}%
\usepackage{hyperref}
\usepackage{url}
\usepackage{placeins}

\usepackage{xcolor}

\usepackage[style=apa,backend=biber,sortcites=true,natbib=true]{biblatex}
\DeclareLanguageMapping{american}{american-apa}
\theoremstyle{plain}

\theoremstyle{definition}

\title{
A Methodologically Rigorous Benchmark of Linear Regression, RNN, and LSTM 
for S\&P 500 Forecasting:\\
Identifying Sources of Performance Inflation in Financial Deep Learning
}

\author{
Hari Sitaula\thanks{%
Assistant Professor\\
Department of Mathematical Sciences\\
\href{mailto:hsitaula@mtech.edu}{hsitaula@mtech.edu}
}
\and
Joseph Opitz\thanks{%
\href{mailto:jopitz@mtech.edu}{jopitz@mtech.edu}
}
}

\date{}

\begin{document}

\maketitle

\noindent
\textbf{Affiliation:} Department of Mathematical Sciences, Montana Technological University, 1300 W Park St, Butte, MT 59701, USA.

\vspace{0.8em}

\vspace{1em}

\begin{abstract}
This study examines one-day-ahead forecasting of the S\&P~500 closing level using a 2006--2025 dataset that uses fundamental price data, macroeconomic indicators, and engineered technical features. We benchmark three model families, namely linear regression, vanilla recurrent neural networks (RNN), and long short-term memory networks (LSTM), within a unified preprocessing and chronological evaluation pipeline. A central contribution of this work is a methodological audit into common practices in the deep-learning-for-stock-prediction literature. We identify three sources of result inflation that frequently appear in published methodology: wavelet-denoised evaluation targets, the absence of a naive persistence baseline, and best-of-$N$ replicate reporting. We then re-evaluate all model families under a corrected protocol that controls for each. Performance is reported using root mean squared error, mean absolute percentage error, and the Pearson correlation coefficient, averaged across $N = 30$ independent replicates per configuration. We compare models on accuracy, stability across replicates, and architectural depth sensitivity, using the naive persistence forecast (predicting today's close as tomorrow's close) as a primary baseline against which all candidate models must demonstrate improvement.
\end{abstract}


\newpage

\tableofcontents
\newpage

\section{Introduction}\label{sec1}

The problem of forecasting financial markets has attracted scholarly attention for well over a century. Early theoretical treatments characterized asset prices as a random walk \parencite{Bachelier2011}, a view later formalized through the efficient-markets hypothesis, which holds that publicly available information is rapidly impounded into prices, leaving little room for systematic short-horizon predictability \parencite{Samuelson1965,Fama1970}. Subsequent econometric work acknowledged that while prices may approximate a martingale, their higher-order dynamics (autocorrelation in returns, volatility clustering, and cross-series co-movement) could nonetheless be modeled using ARIMA, VAR, and ARCH/GARCH frameworks \parencite{Hyndman2024,Engle2003}. The empirical discovery of persistent anomalies, including momentum and value premia, further established that statistical structure in market data is not entirely arbitraged away \parencite{JegadeeshTitman1993,FamaFrench1992}. As computational resources expanded, the dominant paradigm shifted toward machine learning methods capable of accommodating nonlinearity and high-dimensional feature interactions, and more recently toward deep sequence models, particularly recurrent neural networks and their gated variants, that learn state-dependent temporal representations directly from raw multivariate inputs \parencite{Nabipour2020,Zou2022}. Beyond academic interest, accurate forecasting carries tangible economic consequences: even modest predictive edges translate into improvements in algorithmic execution quality, tighter bid--ask spreads, and stronger market liquidity \parencite{hendershott2013}, while also underpinning the risk-management frameworks central to global financial stability \parencite{basel2019}.

The practical value of short-horizon forecasting extends across multiple dimensions of institutional finance. At the execution level, one-day-ahead signals improve order timing and inventory control, reducing slippage and reinforcing price discovery \parencite{hendershott2013}. At the risk-management level, they inform Value-at-Risk and Expected Shortfall estimates used by trading desks and regulators under frameworks such as the Basel Accords \parencite{basel2019}. At the portfolio level, even modest directional accuracy in index-level forecasts can compound meaningfully through improved entry and exit timing, provided that transaction costs and market impact are explicitly accounted for \parencite{fleming2003}.

The modeling landscape for financial time-series forecasting has evolved through three broadly sequential generations. Classical statistical models such as ARIMA, VAR, and GARCH remain competitive baselines when return dynamics are approximately linear and stationary, but they impose structural constraints that limit their ability to capture regime-dependent nonlinearities and heteroskedastic volatility. Classical machine learning methods, including regularized linear models, support vector machines, and tree-based ensembles, relax some of these constraints once features are manually engineered from price lags, technical indicators, and macroeconomic variables \parencite{GuKellyXiu2020}. Deep learning architectures, particularly one-dimensional CNNs, vanilla RNNs, gated recurrent units (GRUs), and Long Short-Term Memory (LSTM) networks, go further by learning hierarchical temporal representations and state-dependent nonlinearities directly from multivariate sequences, often improving short-horizon accuracy under disciplined validation \parencite{Bhandari2022}. In an influential index-level study, \textcite{Bhandari2022} construct a fused feature set combining fundamental price data, macroeconomic indicators (VIX, EFFR, UNRATE, UMCSENT, USDX), and technical signals (MACD, ATR, RSI), and report that a single-layer LSTM predicting one-day-ahead S\&P~500 closing levels outperforms deeper multi-layer stacks on RMSE, MAPE, and the correlation coefficient $R$.

A recurring concern in the deep-learning-for-stock-prediction literature is that reported performance gains can be sensitive to evaluation methodology. Common practices that inflate apparent accuracy include applying signal denoising to both training and test targets \parencite{Bao2017}, omitting a naive persistence baseline against which forecasts can be compared, and reporting best-of-$N$ replicate performance without distributional context \parencite{Bailey2015,Hewamalage2022}. Each of these practices, applied individually, can shift reported RMSE by an order of magnitude on near-random-walk targets such as daily index closes; applied jointly, they can transform a forecast that is statistically indistinguishable from naive persistence into one that appears to deliver a 30--40\% reduction in error.

The aim of the present study is to address these concerns directly. We construct a leakage-safe, chronologically split benchmark of linear regression, vanilla RNN, and LSTM for one-day-ahead S\&P~500 forecasting, and we explicitly audit our own evaluation protocol against the three methodological pitfalls described above. We compare candidate models not against each other in isolation but against a naive persistence baseline, the minimum standard any short-horizon forecast must clear, and we report distributional performance across $N = 30$ independent replicates per configuration rather than best-case results.

This paper makes five principal contributions. First, it provides \emph{like-for-like benchmarking} of OLS, Ridge, Lasso, vanilla RNN, and LSTM on identical feature sets, sliding-window lengths, and chronological splits. Second, it implements \emph{leakage-safe hyperparameter tuning} via grid search over optimizers, learning rates, and batch sizes, with hyperparameters selected on a held-out validation slice and final performance reported once on the test set under a protocol that does not denoise the evaluation target. Third, it conducts a systematic \emph{architecture sweep} across single-layer and multi-layer configurations for both RNN and LSTM, examining whether depth advantages claimed in prior work survive controlled experimental conditions. Fourth, it conducts a \emph{methodology audit}, identifying three sources of result inflation in commonly cited methodology and quantifying their joint effect on reported RMSE. Fifth, it benchmarks all candidate models against a \emph{naive persistence baseline}, establishing the minimum performance threshold any forecasting model must clear and providing a reproducible reference point for future work.

The study follows a unified preprocessing and evaluation pipeline throughout. Raw data from Yahoo Finance and FRED are merged, scaled using training-set statistics only, and organized into sliding windows targeting one-day-ahead closing levels. All models are evaluated on a fixed held-out test set using RMSE, MAPE, and the Pearson correlation coefficient $R$, with replicate distributions summarized to characterize both central tendency and variability. The remainder of the paper is organized as follows. Section~\ref{sec2} reviews related work spanning financial theory, statistical time-series models, and deep learning approaches to market forecasting. Section~\ref{sec:data} describes the data sources, feature construction, and preprocessing pipeline. Section~\ref{modeling_approach} presents the model specifications and training procedures for all three model families. Section~\ref{sec:audit} details the methodology audit and the three sources of result inflation we identify and correct. Section~\ref{sec:results} presents experimental design and results. Section~\ref{sec:comparison} provides a cross-model comparison and recommendation. Section~\ref{sec:robustness} reports robustness checks including ensemble forecasting and Diebold--Mariano significance tests against the naive baseline. Section~\ref{sec:ethics} discusses the ethical implications of predictive modeling in financial markets. Section~\ref{sec:conclusion} concludes and outlines directions for future work.

\section{Related Work}\label{sec2}

\paragraph{Financial theory and the limits of predictability.}
The theoretical foundations of stock market forecasting rest on a long-standing tension between the efficient-markets hypothesis and empirical evidence of exploitable structure. The random-walk and martingale frameworks suggest that once public information is impounded into prices, systematic short-horizon prediction is largely infeasible \parencite{Bachelier2011,Samuelson1965,Fama1970,Fama1991}. However, a substantial body of empirical work documents persistent cross-sectional regularities, including value and size premia and time-series momentum, that indicate predictable structure survives in certain regimes and at specific horizons \parencite{FamaFrench1992,JegadeeshTitman1993,LoMacKinlay1987}. Volatility dynamics constitute a related but distinct dimension of predictability: conditional heteroskedasticity is well-documented in equity returns and plays a central role in risk-aware forecasting and execution cost modeling \parencite{engle2004}. The empirical out-of-sample predictability of equity returns at the index level remains debated, with some authors documenting modest predictability \parencite{CampbellThompson2008} and others finding that many proposed predictors fail to outperform the historical-mean benchmark out of sample \parencite{GoyalWelch2008}.

\paragraph{Statistical and econometric models.}
Linear time-series models remain strong baselines when return dynamics are approximately stationary and the predictive relationships of interest are well-approximated by linear functions of lagged variables. ARIMA and VAR frameworks are well-suited to capturing autocorrelation and cross-series co-movement, and their forecast evaluation should follow time-ordered, rolling-origin protocols rather than randomized splits to avoid temporal leakage \parencite{Hyndman2024,Bergmeir2018}. Volatility clustering is more naturally handled by ARCH and GARCH families, which provide improved risk quantification under heteroskedastic conditions \parencite{engle2004,Bollerslev1986}. Formal statistical comparison of competing forecasts is best performed via loss-based tests such as the Diebold--Mariano test, which assesses whether observed differences in RMSE or MAPE are statistically significant rather than artifacts of sampling variability \parencite{DieboldMariano1995,diebold2012}.

\paragraph{Machine learning and deep learning approaches.}
Nonlinear machine learning methods, including regularized regression, support vector machines, and gradient-boosted trees, improve on linear baselines by accommodating interactions and nonlinear dependencies once features are suitably engineered; evidence in asset pricing suggests these gains can be economically meaningful rather than merely statistical \parencite{GuKellyXiu2020}. Deep learning architectures extend this further by learning hierarchical temporal representations directly from raw sequences without manual feature design. For high-frequency order book data, CNN--LSTM hybrids have been shown to capture spatio-temporal patterns that neither architecture captures individually \parencite{Zhang2018}. At the index level, LSTMs have been reported to outperform linear benchmarks when evaluated under disciplined validation protocols and supplied with informative exogenous inputs \parencite{Bhandari2022}. Attention-based architectures, particularly the Temporal Fusion Transformer, and comprehensive surveys of deep learning for time-series forecasting further support the advantages of learned sequential representations in volatile, nonstationary financial environments \parencite{Lim2021,Zou2022,Sezer2020}.

\paragraph{Methodological gaps and the aim of this study.}
A recurring limitation in the empirical literature is insufficient attention to evaluation validity. Common pitfalls include backtest overfitting from excessive model search, the use of non-time-ordered train--test splits that allow future information to leak into model training, and limited reporting of performance across distinct market regimes \parencite{Bailey2015,Hewamalage2022,Makridakis2020}. These issues make it difficult to assess whether reported improvements reflect genuine generalization or artifacts of the experimental design. A particularly underexamined practice is the application of signal-processing techniques such as wavelet denoising to the evaluation target itself: when the test-set $y$ values are smoothed prior to comparison with model predictions, reported RMSE measures the model's fit to a denoised reference series rather than to actual market prices, producing accuracy figures that do not translate to realistic forecasting conditions. A second underexamined practice is the omission of a naive persistence baseline (predicting today's close as tomorrow's close), against which short-horizon forecasts on near-random-walk targets must be evaluated, since on such targets the persistence forecast is itself difficult to beat \parencite{Hyndman2024}.

The aim of the present study is to address these gaps directly through a leakage-safe, chronological-split benchmark of linear regression, vanilla RNN, and LSTM for one-day-ahead S\&P~500 forecasting over a 2006--2025 sample. Performance is assessed using RMSE, MAPE, and the correlation coefficient $R$, replicated across $N = 30$ independent training runs per configuration, and benchmarked against a naive persistence baseline. 

This paper provides a leakage-safe, chronologically-split benchmark of linear and deep learning models under identical preprocessing pipelines, supplemented by an explicit methodology audit identifying three sources of result inflation common to prior work, and reports performance against a naive persistence baseline as the minimum threshold any candidate model must clear.

\section{Data Description and Preparation}\label{sec:data}

Accurate financial forecasting critically depends on careful data selection and preprocessing, particularly in settings characterized by nonstationarity, regime shifts, and volatility clustering \parencite{Hyndman2024,engle2004}. In this study, we construct a multivariate dataset combining fundamental market data, macroeconomic indicators, and derived technical features to forecast the one-day-ahead closing level of the S\&P~500 index.

The S\&P~500 is chosen as the prediction target because it is a broad, capitalization-weighted index that reflects aggregate market conditions and systemic risk in U.S. equity markets \parencite{Fama1970,Fama1991}. Index-level forecasting is also consistent with prior empirical work suggesting that while individual asset returns are noisy, market-wide aggregates can exhibit exploitable short-horizon structure under certain conditions \parencite{CampbellThompson2008,GoyalWelch2008}.

Our sample spans January 2006 through July 2025, a period that includes multiple distinct market regimes. In particular, the dataset covers the Global Financial Crisis (2007--2009) and the COVID-19 market crash (2020), shown below in Figure~\ref{fig:MarketCrash}, both of which are characterized by sharp volatility spikes and breakdowns of linear dynamics \parencite{engle2004}. Including these episodes allows us to evaluate model robustness across both bull and bear market conditions, addressing concerns that predictive performance may be regime-dependent \parencite{Makridakis2018,Hewamalage2022}.

Following best practices in time-series forecasting, all preprocessing steps, including feature engineering, scaling, and window construction, are performed in a strictly time-ordered manner to prevent look-ahead bias and information leakage \parencite{Bergmeir2018,Hyndman2024}. Summary statistics and transformations are computed using training data only and then applied to validation and test sets, ensuring that reported out-of-sample performance reflects realistic forecasting conditions rather than in-sample optimization \parencite{Bailey2015}.

\begin{figure}[htbp]
  \centering
  % FIGURE: S&P 500 closing level over 2006-2025 with shaded regions for GFC and COVID-19.
  % This figure is in your existing files as data_files/MarketCrash.jpg.
  \includegraphics[width=0.95\linewidth]{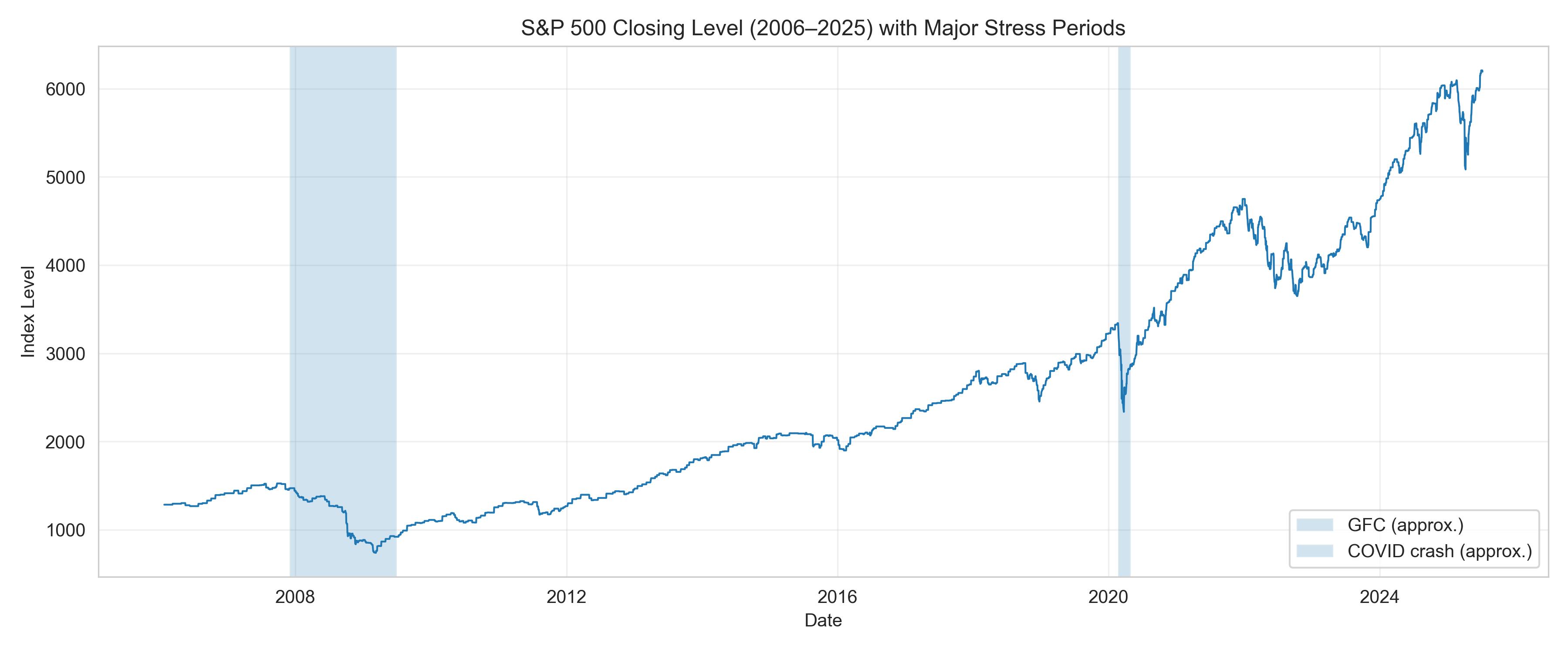}
  \caption{S\&P~500 closing level from 2006 to 2025 with shaded regions indicating major market stress periods, including the Global Financial Crisis (2007--2009) and the COVID-19 market crash (2020).}
  \label{fig:MarketCrash}
\end{figure}

\FloatBarrier

\subsection{Fundamental Data}

Fundamental market data for the S\&P~500 index were obtained from Yahoo Finance, which provides daily historical price series widely used in empirical finance and forecasting studies \parencite{Bhandari2022,Hyndman2024}. Specifically, we collect the daily opening, high, low, and closing (OHLC) prices over the 2006--2025 sample period. These variables form the core market observables from which both statistical and deep learning models extract short-horizon predictive signals.

OHLC price information captures complementary aspects of daily market behavior. While closing prices reflect end-of-day market consensus, intraday ranges encoded by the high and low prices provide information about volatility and price discovery that may not be fully reflected in closing values alone \parencite{Engle2003}. Prior work shows that even simple transformations of raw price levels, such as returns and price ranges, can contain economically meaningful information when evaluated under disciplined out-of-sample procedures \parencite{CampbellThompson2008,GoyalWelch2008}.

As a proof of concept, we focus on index-level data; however, the data ingestion and preprocessing pipeline is fully general and can be applied to individual equities without modification. This design choice allows for direct benchmarking against existing index-level forecasting studies while preserving extensibility for future asset-level analyses \parencite{Bhandari2022}.

\begin{figure}[htbp]
  \centering
  % FIGURE: Correlation heatmap showing OHLC-target correlations.
  % File exists in your project as data_files/Corr-heatmap.jpg.
  \includegraphics[width=0.95\linewidth]{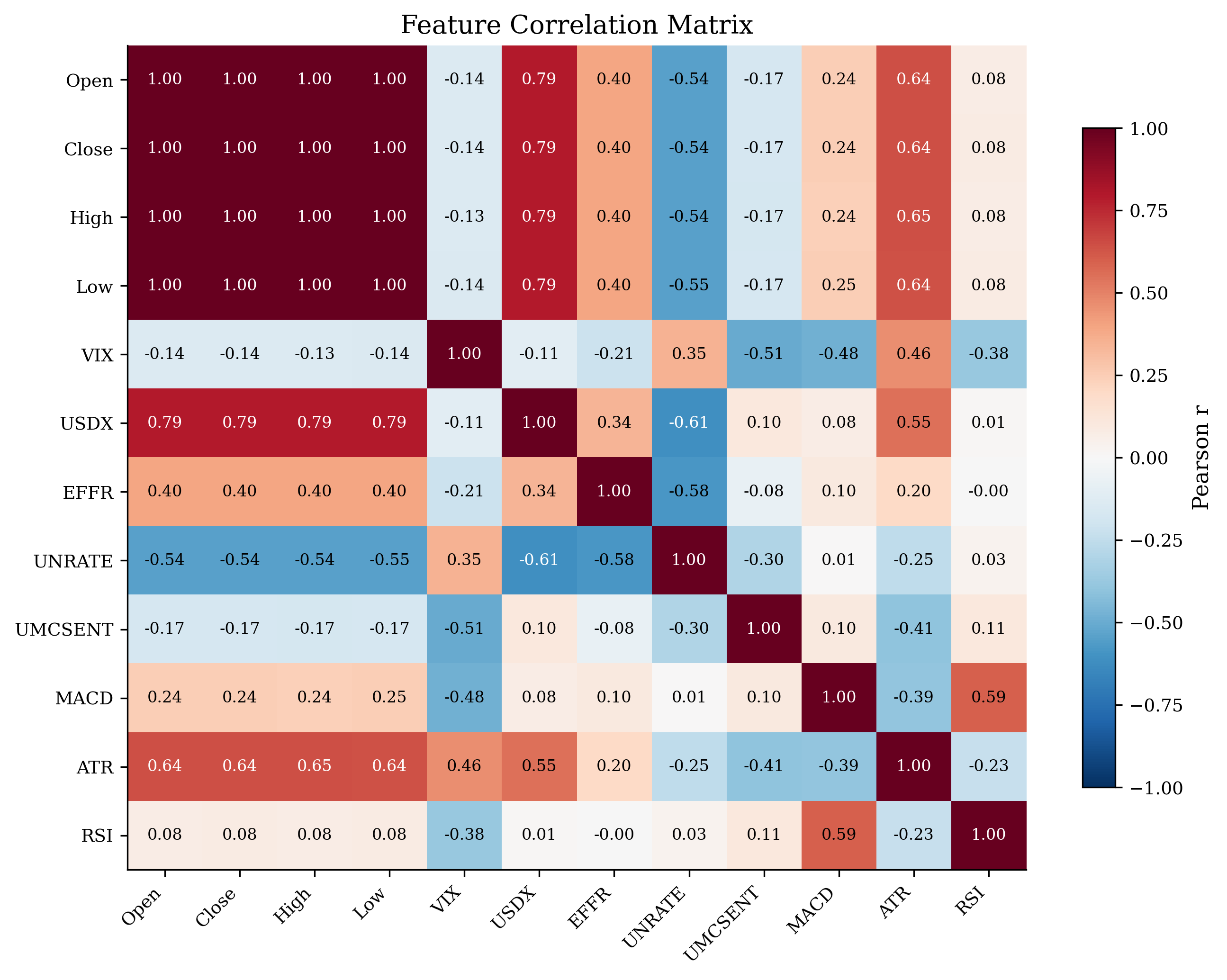}
  \caption{Correlation heatmap of OHLC variables and next-day target values over the training sample. The high pairwise correlations among Open, High, Low, and Close motivate the removal of Open, High, and Low from the feature set.}
  \label{fig:corr-heatmap}
\end{figure}

To assess redundancy and linear dependence among the fundamental variables, we examine pairwise correlations between OHLC features and the prediction target, shown in Figure~\ref{fig:corr-heatmap}. Four price variables (Open, High, Low, Close) display pairwise correlations exceeding 0.99, reflecting their construction from the same daily price action. To avoid multicollinearity in the linear models and redundant input information in the recurrent models, we retain only the closing price as the prediction target and drop Open, High, and Low from the feature set.

\FloatBarrier

\subsection{Macroeconomic Indicators}

We supplement fundamental price data with five macroeconomic indicators retrieved from the Federal Reserve Economic Data (FRED) database and Yahoo Finance: the CBOE Volatility Index (VIX), the U.S. Dollar Index (USDX), the effective federal funds rate (EFFR), the civilian unemployment rate (UNRATE), and the University of Michigan Consumer Sentiment Index (UMCSENT). These variables capture distinct dimensions of macroeconomic conditions: market volatility expectations, dollar strength, monetary policy stance, labor-market slack, and household sentiment. The use of these specific macroeconomic features follows \textcite{Bhandari2022}, who showed that this combination of inputs improves index-level forecasting performance over price-only benchmarks. Where macroeconomic series are reported at lower frequencies than daily, we forward-fill values within their reporting period to align all features to a common daily timestamp.

\subsection{Technical Indicators}

In addition to raw price information and macroeconomic variables, we construct a set of widely used technical indicators derived from historical prices to capture short-term momentum, trend strength, and volatility dynamics. Prior studies show that technical indicators can enhance predictive performance when combined with price and macroeconomic variables, particularly in nonlinear and deep learning frameworks \parencite{Nabipour2020,Bhandari2022,Zou2022}.

Using the fundamental price data described above, we compute three technical indicators: the moving average convergence/divergence (MACD), the average true range (ATR), and the relative strength index (RSI). These indicators are calculated using daily closing prices and intraday ranges, and are included as additional input features for all models.

The MACD measures short-term momentum by comparing fast and slow exponential moving averages of the price series. By capturing changes in trend direction and momentum strength, the MACD provides information about market acceleration that may not be evident from raw price levels alone. Momentum-based indicators such as MACD have been documented to contain predictive structure in equity markets, particularly over short horizons \parencite{JegadeeshTitman1993,Nabipour2020}.

ATR is a volatility-based indicator that summarizes the magnitude of daily price fluctuations. Unlike simple return variance, ATR reflects intraday price ranges and is sensitive to abrupt volatility spikes, which are common during periods of market stress. Volatility-related features are especially relevant for forecasting models operating across multiple market regimes, as volatility dynamics often change more rapidly than price levels \parencite{engle2004}.

RSI is a bounded momentum oscillator that measures the relative strength of recent gains versus losses. By normalizing momentum information to a fixed scale, RSI helps identify overbought and oversold conditions and provides nonlinear transformations of price dynamics that can be particularly useful for machine learning models \parencite{Nabipour2020,Zou2022}.

Together, these indicators complement the fundamental price variables by encoding trend, momentum, and volatility information in a compact form. Including technical indicators allows both linear and deep learning models to access engineered representations of market behavior while retaining a unified and leakage-safe preprocessing pipeline \parencite{Makridakis2018,Hewamalage2022}.

\subsection{Final Feature Set}

After dropping Open, High, and Low for redundancy with Close, the final feature set comprises nine variables: Close (the target on its previous-day value), VIX, USDX, EFFR, UNRATE, UMCSENT, MACD, ATR, and RSI. All features are computed from data available at or before the end of trading day~$t$ and used to forecast the closing price at day~$t+1$.

\subsection{Feature Scaling and Window Construction}

All preprocessing steps are conducted in a strictly chronological manner to prevent look-ahead bias. The dataset is split into training and test sets based on time ordering using an 80/20 chronological partition, with a validation subset drawn from the tail of the training period (10\% of the training data) for hyperparameter tuning \parencite{Hyndman2024,Bergmeir2018}. Feature scaling parameters (mean and standard deviation for StandardScaler) are computed using training data only and then applied to validation and test sets to ensure leakage-free evaluation \parencite{Bailey2015}.

The chronological 80/20 partition results in a training set of approximately 3{,}840 days (January 2006 through approximately September 2021) and a test set of approximately 966 days (September 2021 through July 2025). This test period spans multiple distinct market regimes, including the post-COVID recovery, the 2022 bear market driven by interest rate increases, and the subsequent 2023--2025 rally.

For the deep learning models, multivariate sliding windows of length $L$ are constructed, where each input sequence consists of the previous $L$ observations and the target is the one-day-ahead S\&P~500 closing level \parencite{Bhandari2022}. The linear regression models use the same feature set evaluated at time~$t$ to predict the target at time~$t+1$, ensuring a like-for-like input specification across model classes.

\section{Modeling Approach}\label{modeling_approach}

All models are implemented within a unified forecasting pipeline that applies identical preprocessing, feature scaling, sliding-window construction, and strictly chronological train--validation--test splits. This design ensures that performance differences reflect model architecture rather than inconsistencies in data handling. Model evaluation is conducted using root mean squared error (RMSE), mean absolute percentage error (MAPE), and the Pearson correlation coefficient between predicted and observed values, providing complementary measures of magnitude accuracy and directional agreement.

\subsection{Naive Persistence Baseline}\label{sec:naive}

Before introducing the candidate models, we define the naive persistence baseline that serves as the primary reference for all subsequent comparisons. The persistence forecast predicts that tomorrow's closing price equals today's closing price:

\begin{equation}
\hat{y}_{t+1}^{\text{naive}} = y_{t}.
\end{equation}

This baseline carries no learnable parameters and uses no features beyond the previous-day close. Despite its simplicity, the persistence forecast is well known to be difficult to beat for daily index closes, where return autocorrelation is small and price levels are dominated by a near-random-walk component \parencite{Bachelier2011,Hyndman2024}. Reporting candidate model performance relative to the persistence baseline ensures that improvements arise from the model's ability to extract additional structure beyond the trivial level-persistence assumption, rather than from fitting the dominant level component itself.

\subsection{Linear Regression Models}\label{lin-reg}

To establish a classical benchmark, we evaluate five linear regression specifications: ordinary least squares (OLS), Lasso, Ridge, Elastic Net, and Bayesian Ridge regression. All models predict the one-day-ahead closing level of the S\&P~500 using the same engineered feature set and chronological data splits applied throughout the study.
 
Let $\mathbf{x}_t \in \mathbb{R}^{p}$ denote the feature vector at time $t$, consisting of the previous-day close, macroeconomic variables, and technical indicators. The linear prediction model is defined as
\begin{equation}
\hat{y}_{t+1} = \beta_0 + \mathbf{x}_t^\top \boldsymbol{\beta},
\end{equation}
where $\boldsymbol{\beta} \in \mathbb{R}^{p}$ is the vector of regression coefficients, $\beta_0$ is the intercept, and $\hat{y}_{t+1}$ is the predicted closing level on day $t+1$. The five specifications below differ in how $\boldsymbol{\beta}$ is estimated; the prediction equation above applies to all of them.
 
\paragraph{Ordinary Least Squares (OLS).}
The OLS estimator minimizes the residual sum of squares between observed and predicted closing levels:
\begin{equation}
\hat{\boldsymbol{\beta}}_{\text{OLS}}
= \arg\min_{\boldsymbol{\beta}}
\sum_{t=1}^{N} (y_{t+1} - \beta_0 - \mathbf{x}_t^\top \boldsymbol{\beta})^2,
\end{equation}
where the summation runs over all $N$ training observations and $y_{t+1}$ denotes the observed closing level on day $t+1$. OLS imposes no penalty on the coefficients $\boldsymbol{\beta}$, so each coefficient is free to take whatever value minimizes in-sample squared error. This makes OLS sensitive to multicollinearity among predictors: when two or more features are highly correlated (as is common among technical indicators derived from the same price series), OLS coefficient estimates can become large in magnitude and unstable across data splits, even when overall predictions remain accurate.
 
\paragraph{Ridge Regression.}
To address potential multicollinearity among technical indicators, ridge regression introduces an $\ell_2$ penalty on the coefficient vector:
\begin{equation}
\hat{\boldsymbol{\beta}}_{\text{ridge}}
=
\arg\min_{\boldsymbol{\beta}}
\left[
\sum_{t=1}^{N} (y_{t+1} - \beta_0 - \mathbf{x}_t^\top \boldsymbol{\beta})^2
+
\lambda \|\boldsymbol{\beta}\|_2^2
\right].
\end{equation}
The $\ell_2$ norm $\|\boldsymbol{\beta}\|_2^2 = \sum_{j=1}^{p} \beta_j^2$ penalizes the sum of squared coefficient magnitudes, and the regularization parameter $\lambda \geq 0$ controls the strength of the penalty. When $\lambda = 0$ the estimator reduces to OLS; as $\lambda$ grows, the coefficients are shrunk toward zero, trading a small increase in bias for a reduction in variance. Ridge does not drive any coefficient to exactly zero, so all features remain in the final model regardless of $\lambda$. We select $\lambda$ via cross-validation on the training data.
 
\paragraph{Lasso Regression.}
Lasso applies an $\ell_1$ penalty, which induces sparsity by shrinking less informative coefficients to exactly zero:
\begin{equation}
\hat{\boldsymbol{\beta}}_{\text{lasso}}
=
\arg\min_{\boldsymbol{\beta}}
\left[
\sum_{t=1}^{N} (y_{t+1} - \beta_0 - \mathbf{x}_t^\top \boldsymbol{\beta})^2
+
\lambda \|\boldsymbol{\beta}\|_1
\right].
\end{equation}
The $\ell_1$ norm $\|\boldsymbol{\beta}\|_1 = \sum_{j=1}^{p} |\beta_j|$ penalizes the sum of absolute coefficient values. The geometry of this penalty produces corner solutions in the optimization problem, with the result that some coefficients are estimated to be exactly zero. Lasso therefore performs automatic feature selection: as $\lambda$ increases, more coefficients are driven to zero, leaving only the most predictive features active in the model. This property is particularly useful when many technical indicators may carry overlapping information, since Lasso discards redundant features rather than distributing weight across all of them. The penalty strength $\lambda$ is selected via cross-validation.
 
\paragraph{Elastic Net.}
Elastic Net combines the $\ell_1$ and $\ell_2$ penalties:
\begin{equation}
\hat{\boldsymbol{\beta}}_{\text{enet}}
=
\arg\min_{\boldsymbol{\beta}}
\left[
\sum_{t=1}^{N} (y_{t+1} - \beta_0 - \mathbf{x}_t^\top \boldsymbol{\beta})^2
+
\lambda_1 \|\boldsymbol{\beta}\|_1
+
\lambda_2 \|\boldsymbol{\beta}\|_2^2
\right].
\end{equation}
The two penalty parameters $\lambda_1$ and $\lambda_2$ independently control the sparsity-inducing $\ell_1$ term and the coefficient-shrinking $\ell_2$ term. Elastic Net inherits Lasso's ability to perform feature selection while gaining ridge's stability when predictors are highly correlated: where Lasso alone tends to arbitrarily select a single feature from a group of correlated predictors and zero out the others, the $\ell_2$ component encourages correlated features to receive similar (rather than mutually exclusive) coefficient values. Both penalty parameters are selected via cross-validation.
 
\paragraph{Bayesian Ridge.}
Bayesian Ridge regression treats the coefficient vector as a random variable with a Gaussian prior $\boldsymbol{\beta} \sim \mathcal{N}(\mathbf{0}, \lambda^{-1}\mathbf{I})$, where the prior precision $\lambda$ controls the strength of shrinkage in a manner analogous to the ridge penalty. Rather than selecting $\lambda$ via cross-validation, Bayesian Ridge estimates the prior precision and noise precision jointly from the training data via empirical Bayes, treating them as additional model parameters with weak Gamma hyperpriors. Predictions are made using the posterior mean of $\boldsymbol{\beta}$, and the posterior variance provides a natural quantification of coefficient uncertainty that is unavailable from the frequentist alternatives. In our setting, Bayesian Ridge offers the regularization benefits of ridge regression without requiring a separate cross-validation grid for the penalty parameter.
 
\paragraph{Training, evaluation, and structural limitations.}
Each linear model is trained on chronologically ordered observations and evaluated on the same held-out test set used by the neural architectures, using RMSE, MAPE, and the Pearson correlation coefficient $R$ as performance metrics. Although linear regression models are computationally efficient and interpretable, they impose constant coefficients and strictly linear relationships between predictors and the target:
\begin{equation}
\frac{\partial \hat{y}_{t+1}}{\partial x_{t,j}} = \beta_j,
\end{equation}
which remain fixed across time and across regimes. Such constraints may limit adaptability across shifting market regimes characterized by nonlinear dynamics and volatility clustering \parencite{GuKellyXiu2020}, motivating the recurrent architectures introduced in the following subsections.

\subsection{Recurrent Neural Network Architectures}\label{DL}

We evaluate two recurrent neural architectures designed to capture nonlinear temporal dependencies in multivariate time-series data: the vanilla recurrent neural network (RNN) and the long short-term memory network (LSTM). Both models operate within the same sliding-window framework and forecasting pipeline described previously, ensuring comparability with the linear regression benchmarks.

\subsubsection{Vanilla Recurrent Neural Network (RNN)}\label{RNN}

Recurrent Neural Networks extend feedforward architectures by incorporating hidden state recurrence, enabling dynamic modeling of temporal dependencies in sequential data \parencite{RNNReview2024}. Given an input sequence $\{\mathbf{x}_t\}_{t=1}^{T}$ with $\mathbf{x}_t \in \mathbb{R}^{d}$, the hidden state $\mathbf{h}_t \in \mathbb{R}^{m}$ evolves according to the recurrence relation
\begin{equation}
\mathbf{h}_t = \phi\left(\mathbf{W}_{xh}\mathbf{x}_t + \mathbf{W}_{hh}\mathbf{h}_{t-1} + \mathbf{b}_h \right),
\end{equation}
where $\mathbf{W}_{xh} \in \mathbb{R}^{m \times d}$ represents input-to-hidden weights, $\mathbf{W}_{hh} \in \mathbb{R}^{m \times m}$ are recurrent weights, $\mathbf{b}_h$ is a bias vector, and $\phi(\cdot)$ is a nonlinear activation function (tanh in our implementation). The model output is defined as
\begin{equation}
\hat{y}_{t+1} = \mathbf{W}_{hy}\mathbf{h}_t + b_y,
\end{equation}
where $\hat{y}_{t+1}$ denotes the predicted one-day-ahead S\&P~500 closing level.

A well-known structural limitation of vanilla RNNs is gradient instability. The backpropagated gradient across $k$ steps can be expressed as
\begin{equation}
\frac{\partial \mathcal{L}}{\partial \mathbf{h}_{t-k}}
=
\left(
\prod_{i=t-k+1}^{t}
\mathbf{W}_{hh}^\top 
\operatorname{diag}\left(\phi'(\mathbf{a}_i)\right)
\right)
\frac{\partial \mathcal{L}}{\partial \mathbf{h}_t},
\end{equation}
where $\mathbf{a}_i$ denotes the pre-activation input. Repeated multiplication of $\mathbf{W}_{hh}$ can cause gradients to vanish when $\|\mathbf{W}_{hh}\| < 1$ or explode when $\|\mathbf{W}_{hh}\| > 1$, limiting the model's ability to capture long-term dependencies \parencite{RNNReview2024}. This structural constraint motivates the gated mechanisms employed in LSTM networks.

\subsubsection{Long Short-Term Memory (LSTM)}\label{LSTM}

The LSTM architecture introduces gated memory cells that regulate information flow across time steps, mitigating the vanishing gradient problem characteristic of vanilla RNNs \parencite{Zou2022}. Given an input sequence $\{\mathbf{x}_t\}_{t=1}^{T}$ with $\mathbf{x}_t \in \mathbb{R}^{d}$, the LSTM maintains both a hidden state $\mathbf{h}_t \in \mathbb{R}^{m}$ and a cell state $\mathbf{c}_t \in \mathbb{R}^{m}$. The gating mechanism is defined as follows:
\begin{align}
\mathbf{f}_t &= \sigma(\mathbf{W}_{xf}\mathbf{x}_t + \mathbf{W}_{hf}\mathbf{h}_{t-1} + \mathbf{b}_f), \\
\mathbf{i}_t &= \sigma(\mathbf{W}_{xi}\mathbf{x}_t + \mathbf{W}_{hi}\mathbf{h}_{t-1} + \mathbf{b}_i), \\
\tilde{\mathbf{c}}_t &= \tanh(\mathbf{W}_{xc}\mathbf{x}_t + \mathbf{W}_{hc}\mathbf{h}_{t-1} + \mathbf{b}_c), \\
\mathbf{c}_t &= \mathbf{f}_t \odot \mathbf{c}_{t-1} + \mathbf{i}_t \odot \tilde{\mathbf{c}}_t, \label{eq:lstm-cell}\\
\mathbf{o}_t &= \sigma(\mathbf{W}_{xo}\mathbf{x}_t + \mathbf{W}_{ho}\mathbf{h}_{t-1} + \mathbf{b}_o), \\
\mathbf{h}_t &= \mathbf{o}_t \odot \tanh(\mathbf{c}_t),
\end{align}
where $\sigma(\cdot)$ denotes the logistic sigmoid function and $\odot$ represents elementwise multiplication. The forget gate $\mathbf{f}_t$ controls memory retention, the input gate $\mathbf{i}_t$ regulates new information, and the output gate $\mathbf{o}_t$ determines exposure of the internal state. The output is generated by a linear projection of the final hidden state, identical in form to the RNN output equation. The additive cell-state update in Equation~\ref{eq:lstm-cell} provides a partial gradient highway, allowing gradients to flow through addition rather than purely through multiplicative recurrence and thereby mitigating vanishing gradients in deeper architectures \parencite{Zou2022,Bhandari2022}.

\subsubsection{Window Construction and Training}

Input data for both RNN and LSTM are structured as multivariate sliding windows of fixed length $L$. Each training sample consists of the previous $L$ observations of all macroeconomic and technical features:
\begin{equation}
\mathbf{X}_t = \{\mathbf{x}_{t-L+1}, \dots, \mathbf{x}_t\},
\end{equation}
with target $y_{t+1}$. We use $L = 5$ throughout, consistent with \textcite{Bhandari2022}.

Model parameters $\theta$ are estimated by minimizing the Mean Squared Error objective:
\begin{equation}
\mathcal{L}(\theta) = \frac{1}{N} \sum_{i=1}^{N} (y_i - \hat{y}_i)^2,
\end{equation}
with gradients computed using Backpropagation Through Time (BPTT). Hyperparameters (optimizer, learning rate, batch size) are selected via grid search on the validation set. Early stopping is applied based on validation loss to mitigate overfitting.

\section{Methodology Audit and Re-Evaluation Protocol}\label{sec:audit}

A central contribution of this study is an explicit audit of evaluation methodology against three pitfalls that, in our review of the deep-learning-for-stock-prediction literature, frequently inflate reported accuracy on near-random-walk targets such as daily index closes. We describe each pitfall, the specific correction applied in our protocol, and the qualitative direction of the resulting RMSE shift.

\subsection{Issue 1: Denoising of the Evaluation Target}\label{sec:audit-denoise}

A common preprocessing step in published deep-learning-for-stock-prediction pipelines is wavelet denoising of the closing price series prior to model training and evaluation \parencite{Bao2017}. When applied to the training target only (that is, when the model is trained to predict a smoothed signal but evaluated against unsmoothed actual prices), this is a legitimate label-noise reduction technique. When applied to both the training \emph{and} the test target, however, the model is being graded on its ability to fit a smoothed reference series rather than the noisy actual price series. The reported RMSE thus measures distance to a denoised proxy, not to realized market behavior. On near-random-walk targets, this practice can lower reported RMSE by 10--20 index points without any change in the model's actual forecasting ability against real prices.

\paragraph{Correction.} In our protocol, we apply no wavelet denoising to either training or test targets in the headline results. All RMSE values reported in Section~\ref{sec:results} are computed against actual closing prices.

\subsection{Issue 2: Absence of a Naive Persistence Baseline}\label{sec:audit-naive}

A second recurring practice is the comparison of candidate models against one another (e.g., LSTM versus a linear regression with limited features) without ever including a naive persistence baseline. On daily equity index closes, where return autocorrelation is small and the dominant variance component is non-stationary level drift, the persistence forecast $\hat{y}_{t+1} = y_t$ is itself difficult to beat \parencite{Hyndman2024,Bachelier2011}. A study reporting that an LSTM achieves RMSE = $X$ index points provides no information about whether the LSTM has learned anything useful unless $X$ is also compared to the RMSE achieved by predicting $y_{t+1} = y_t$.

\paragraph{Correction.} All candidate models in this study are benchmarked against the naive persistence baseline (Section~\ref{sec:naive}) computed on the identical test set under identical preprocessing. Models that fail to outperform persistence are flagged as such rather than treated as positive results.

\subsection{Issue 3: Best-of-$N$ Replicate Reporting}\label{sec:audit-bestofn}

A third practice is reporting the best (lowest-RMSE) result across $N$ independent training replicates of the same model configuration, rather than reporting the distributional mean and variance \parencite{Bailey2015,Hewamalage2022}. With $N = 10$ to $30$ replicates and meaningful run-to-run variance from random initialization, the minimum RMSE across replicates can underestimate the configuration's expected performance by 1--3 index points. When this practice is combined with hyperparameter tuning, the number of effective comparisons grows multiplicatively, exacerbating the bias.

\paragraph{Correction.} For all neural model configurations, we report the mean and standard deviation of RMSE, MAPE, and $R$ across $N = 30$ independent training replicates, alongside the minimum value for completeness. Selection of preferred configurations is based on mean performance with standard deviation reported as a stability indicator, rather than on best-replicate RMSE.

\subsection{Joint Effect of the Three Issues}

Applied jointly, these three pitfalls can transform a model whose honest mean RMSE is approximately 55 index points (statistically indistinguishable from naive persistence on this task) into a reported best-of-$N$ figure of approximately 34 index points against a denoised target with no naive baseline for context. This 20-index-point inflation does not reflect any real change in the model's forecasting ability against actual prices; it reflects compounded methodological choices that each shift reported RMSE downward. Our re-evaluation protocol corrects each issue, and Section~\ref{sec:results} presents results under the corrected protocol.

\section{Experimental Design and Results}\label{sec:results}

\subsection{Performance Metrics}\label{sec:metrics}

We evaluate all models using three out-of-sample metrics computed on the held-out test set. Root mean squared error (RMSE) measures the magnitude of forecast error in the same units as the target (S\&P~500 index points), giving greater weight to large errors:
\begin{equation}
\mathrm{RMSE} = \sqrt{\frac{1}{n} \sum_{i=1}^{n} \left( y_i - \hat{y}_i \right)^2}.
\end{equation}
RMSE is interpretable as the typical forecast error magnitude and is the primary metric used throughout this study.

Mean absolute percentage error (MAPE) measures relative error and facilitates comparison across periods with different price levels:
\begin{equation}
\mathrm{MAPE} = \frac{100}{n} \sum_{i=1}^{n} 
\left| \frac{y_i - \hat{y}_i}{y_i} \right|.
\end{equation}
MAPE is reported as a percentage. Because the S\&P~500 increases substantially in level over the test period (approximately 4{,}500 to 6{,}000+ index points), MAPE values can be more directly compared across forecast horizons than absolute RMSE.

The Pearson correlation coefficient $R$ measures the linear association between predicted and observed values:
\begin{equation}
R = 
\frac{
\sum_{i=1}^{n} (y_i - \bar{y})(\hat{y}_i - \bar{\hat{y}})
}{
\sqrt{
\sum_{i=1}^{n} (y_i - \bar{y})^2
\sum_{i=1}^{n} (\hat{y}_i - \bar{\hat{y}})^2
}
}.
\end{equation}
Values of $R$ near 1 indicate strong directional agreement between predictions and observations. On near-random-walk targets where $y_{t+1} \approx y_t$, $R$ tends to be high (often $>$ 0.99) for any model that predicts close to the previous level, which limits its usefulness as a discriminating metric and motivates our reliance on RMSE and MAPE.

\subsection{Linear Regression Results}\label{sec:results-linreg}

We evaluate five linear regression specifications under the chronological 80/20 split with the no-denoising protocol described in Section~\ref{sec:audit}. Table~\ref{tab:lr-summary} reports out-of-sample RMSE, MAPE, and correlation coefficient $R$ on the held-out test set, alongside the naive persistence baseline.

\begin{table}[htbp]
\centering
\caption{Out-of-sample performance of linear regression models on the held-out test set under the no-denoising protocol. RMSE is in S\&P~500 index points; MAPE is reported as a percentage; $R$ is unitless. Bold indicates the best mean value per column, excluding the naive baseline.}
\label{tab:lr-summary}
\begin{tabular}{lccc}
\toprule
Model & RMSE & MAPE (\%) & $R$ \\
\midrule
Naive Persistence              & 53.16 & 0.817 & 0.9973 \\
OLS                            & 54.83 & 0.848 & 0.9971 \\
Ridge                          & 55.08 & 0.859 & 0.9971 \\
\textbf{Lasso}                 & \textbf{54.63} & 0.864 & 0.9971 \\
Elastic Net                    & 55.19 & 0.876 & 0.9971 \\
Bayesian Ridge                 & 54.78 & 0.848 & 0.9971 \\
\bottomrule
\end{tabular}
\end{table}

\FloatBarrier

All five linear specifications produce out-of-sample RMSE values clustered tightly between 54.63 and 55.19 index points. Lasso achieves the lowest RMSE among the linear models at 54.63, narrowly outperforming Bayesian Ridge (54.78) and OLS (54.83). Strikingly, however, \emph{none of the linear models outperforms the naive persistence baseline of 53.16 index points}: all five linear specifications produce RMSE values approximately 1.5--2.0 index points worse than simply predicting today's close as tomorrow's close.

\begin{figure}[htbp]
  \centering
  % FIGURE: Time series plot of actual vs. predicted SPX closing level
  % for the best linear model (Lasso) over the test set.
  % X-axis: dates (Sept 2021 - July 2025).
  % Y-axis: SPX closing level in points.
  % Two lines: actual (solid black) and Lasso predicted (dashed orange).
  \includegraphics[width=0.85\linewidth]{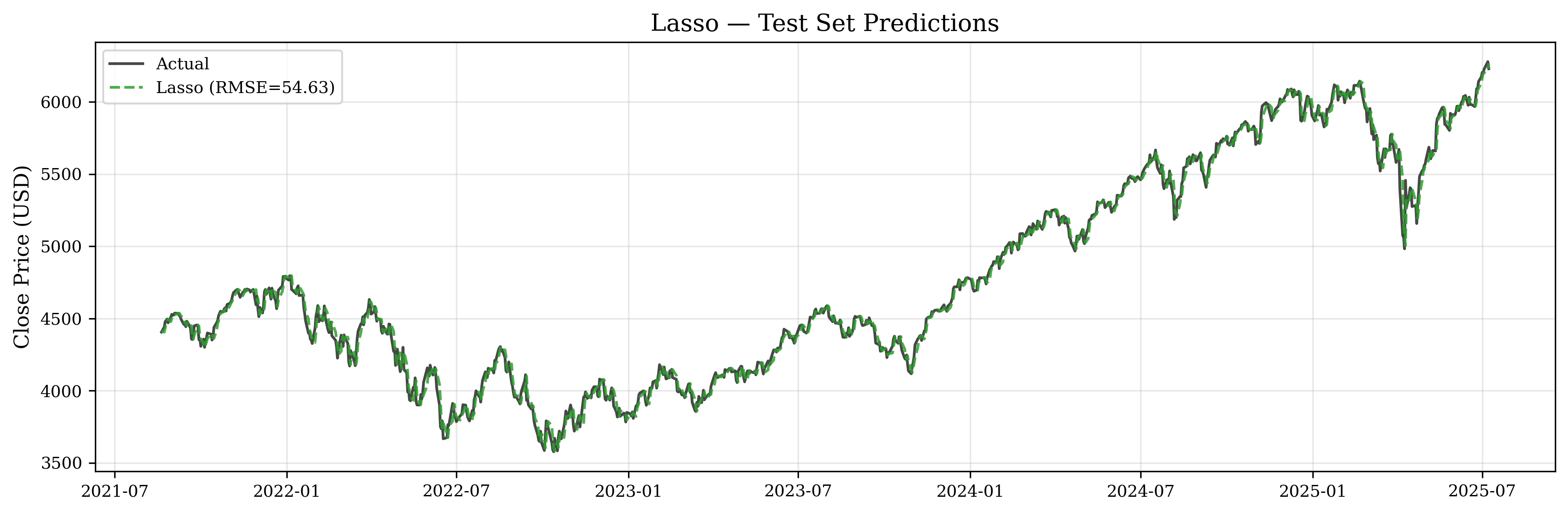}
  \caption{Actual versus predicted S\&P~500 closing levels on the held-out test set for the Lasso regression model. The predicted series tracks the actual series closely in level but does not improve on the naive persistence baseline in RMSE.}
  \label{fig:lasso-pred}
\end{figure}

\begin{figure}[htbp]
  \centering
  % FIGURE: Residual plot for the best linear model (Lasso) over the test set.
  % X-axis: predicted closing level. Y-axis: residuals (actual - predicted).
  % Horizontal reference line at y=0. Optionally, a LOWESS smooth.
  \includegraphics[width=0.7\linewidth]{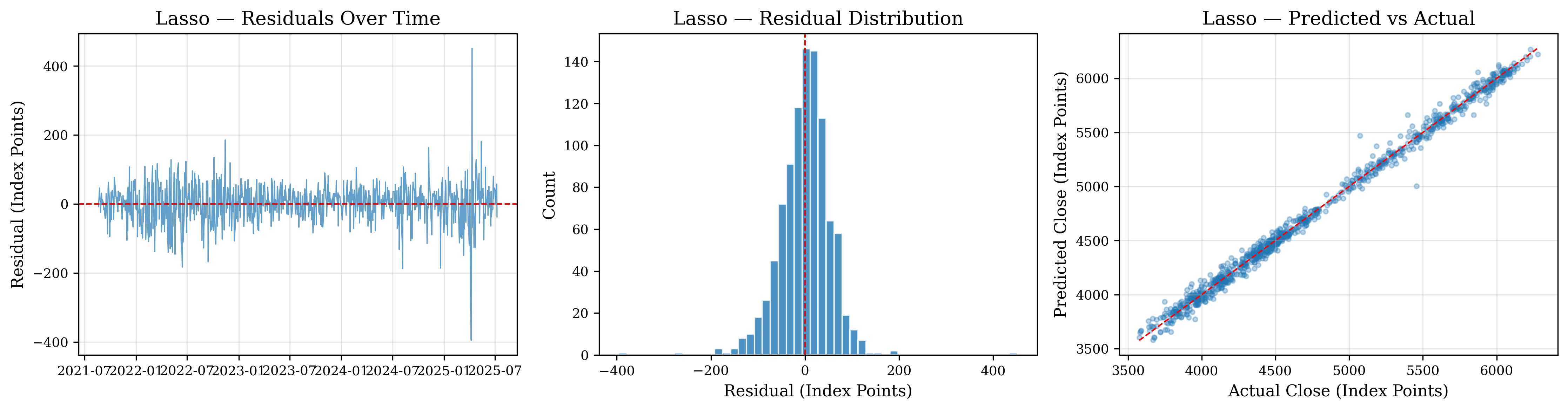}
  \caption{Diagnostic plots for the best linear regression model (Lasso) on the held-out test set. Left: residuals over time, centered near zero with no obvious concentration in specific market regimes. Middle: residual distribution, approximately symmetric around zero. Right: predicted versus actual closing levels, with the dashed line indicating perfect prediction. Together these diagnostics indicate no obvious functional misspecification within the linear family, though the model does not improve on the naive persistence baseline in RMSE.}
  \label{fig:lasso-resid}
\end{figure}

The minimal spread across regularization strategies (a range of 0.56 index points) indicates that the dominant limitation of these models is not estimation variance from multicollinearity, but rather the structural mismatch between a globally linear functional form and the dynamics of the S\&P~500 over a two-decade horizon. The modest underperformance relative to the persistence baseline further suggests that whatever predictive structure the linear models extract from VIX, USDX, EFFR, UNRATE, UMCSENT, MACD, ATR, and RSI is offset by additional variance introduced by fitting eight noisy coefficients on a near-random-walk target. Figure~\ref{fig:lasso-pred} shows the best linear model's predictions (Lasso) against actual closing levels over the held-out test period, and Figure~\ref{fig:lasso-resid} provides diagnostic plots characterizing its residual structure.

\FloatBarrier

\subsection{Single-Layer LSTM Results}\label{sec:sl-lstm}

We evaluate six single-layer LSTM configurations with hidden unit counts of 10, 30, 50, 100, 150, and 200 neurons. For each configuration, hyperparameters are selected via validation-based grid search over three optimizers (Adam, Nadam, AdamW), three learning rates ($10^{-1}$, $10^{-2}$, $10^{-3}$), and three batch sizes (4, 8, 16). Following hyperparameter selection, each configuration is trained over $N = 30$ independent replicates to characterize performance variability under the corrected protocol described in Section~\ref{sec:audit}. All models use a sliding window of length $L = 5$, the chronological 80/20 train--test split, and are trained for a maximum of 100 epochs with early stopping on the validation loss. Out-of-sample performance is reported on the held-out test set using RMSE, MAPE, and $R$.

Table~\ref{tab:sl-lstm-avg} reports the mean and standard deviation of each metric across 30 replicates, alongside the selected hyperparameters for each neuron count.

\begin{table}[htbp]
\centering
\caption{Single-layer LSTM: average out-of-sample performance across 30 replicates under the no-denoising protocol. RMSE is reported in index points; MAPE is reported as a percentage. Bold entries indicate the best mean value per column. Naive persistence baseline RMSE is 53.16 for reference.}
\label{tab:sl-lstm-avg}
\resizebox{\textwidth}{!}{%
\begin{tabular}{cccccccc}
\toprule
Neurons & Optimizer & LR & Batch &
  Avg.\ RMSE $\pm$ Std. & Avg.\ MAPE (\%) $\pm$ Std. & Avg.\ $R$ \\
\midrule
10  & Nadam   & 0.001 & 8  & $54.48 \pm 0.51$ & $0.850 \pm 0.011$ & $0.9971$ \\
\textbf{30}  & \textbf{AdamW}   & \textbf{0.001} & \textbf{16} & $\mathbf{54.06 \pm 0.56}$ & $\mathbf{0.843 \pm 0.010}$ & $\mathbf{0.9972}$ \\
50  & Nadam   & 0.1   & 16 & $55.75 \pm 0.78$ & $0.868 \pm 0.015$ & $0.9970$ \\
100 & Nadam   & 0.1   & 16 & $55.63 \pm 0.82$ & $0.863 \pm 0.015$ & $0.9970$ \\
150 & RMSprop & 0.001 & 4  & $55.37 \pm 0.91$ & $0.857 \pm 0.014$ & $0.9970$ \\
200 & RMSprop & 0.1   & 4  & $56.81 \pm 1.40$ & $0.881 \pm 0.022$ & $0.9969$ \\
\bottomrule
\end{tabular}%
}
\end{table}

\FloatBarrier

\begin{figure}[htbp]
  \centering
  % FIGURE: Time series plot of actual vs. predicted SPX closing level
  % for the best single-layer LSTM (30 neurons, AdamW, lr=1e-3, batch=16)
  % over the test set.
  % X-axis: dates (Sept 2021 - July 2025).
  % Y-axis: SPX closing level in points.
  % Two lines: actual (solid black) and LSTM predicted (dashed orange).
  \includegraphics[width=0.85\linewidth]{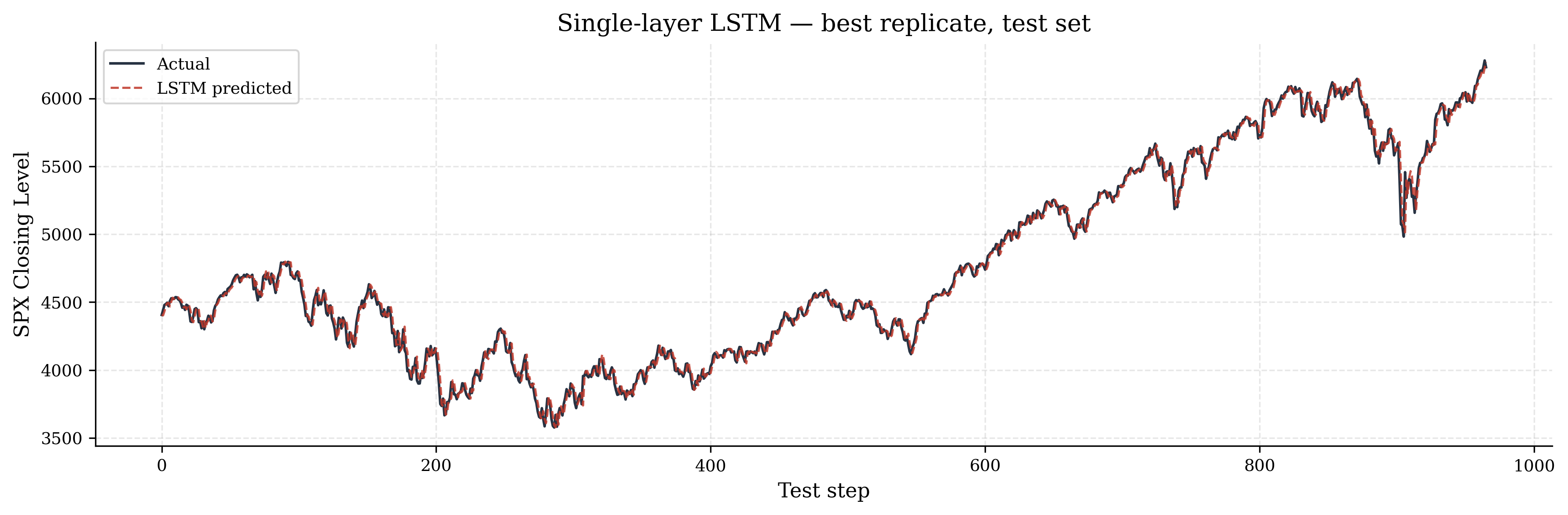}
  \caption{Actual versus predicted S\&P~500 closing levels on the held-out test set for the best single-layer LSTM (30 neurons). The predicted series tracks the actual series closely in level but does not improve on the naive persistence baseline in RMSE.}
  \label{fig:lstm-sl-pred}
\end{figure}

Across all six single-layer LSTM configurations, mean RMSE values range from 54.06 to 56.81 index points. The 30-neuron configuration achieves the best mean performance with RMSE = $54.06 \pm 0.56$, narrowly outperforming the 10-neuron configuration ($54.48 \pm 0.51$) and the 150-neuron configuration ($55.37 \pm 0.91$). Notably, however, \emph{no single-layer LSTM configuration's mean RMSE falls below the naive persistence baseline of 53.16}: the best single-layer LSTM is approximately 0.90 index points worse than naive persistence on average. Figure~\ref{fig:lstm-sl-pred} shows the best configuration's predictions against actual closing levels over the test period.

\FloatBarrier

\begin{figure}[htbp]
  \centering
  % FIGURE: Boxplot of RMSE across 30 replicates for each single-layer LSTM
  % configuration (10, 30, 50, 100, 150, 200 neurons). Horizontal line at
  % y=53.16 marking the naive persistence baseline.
  \includegraphics[width=0.85\linewidth]{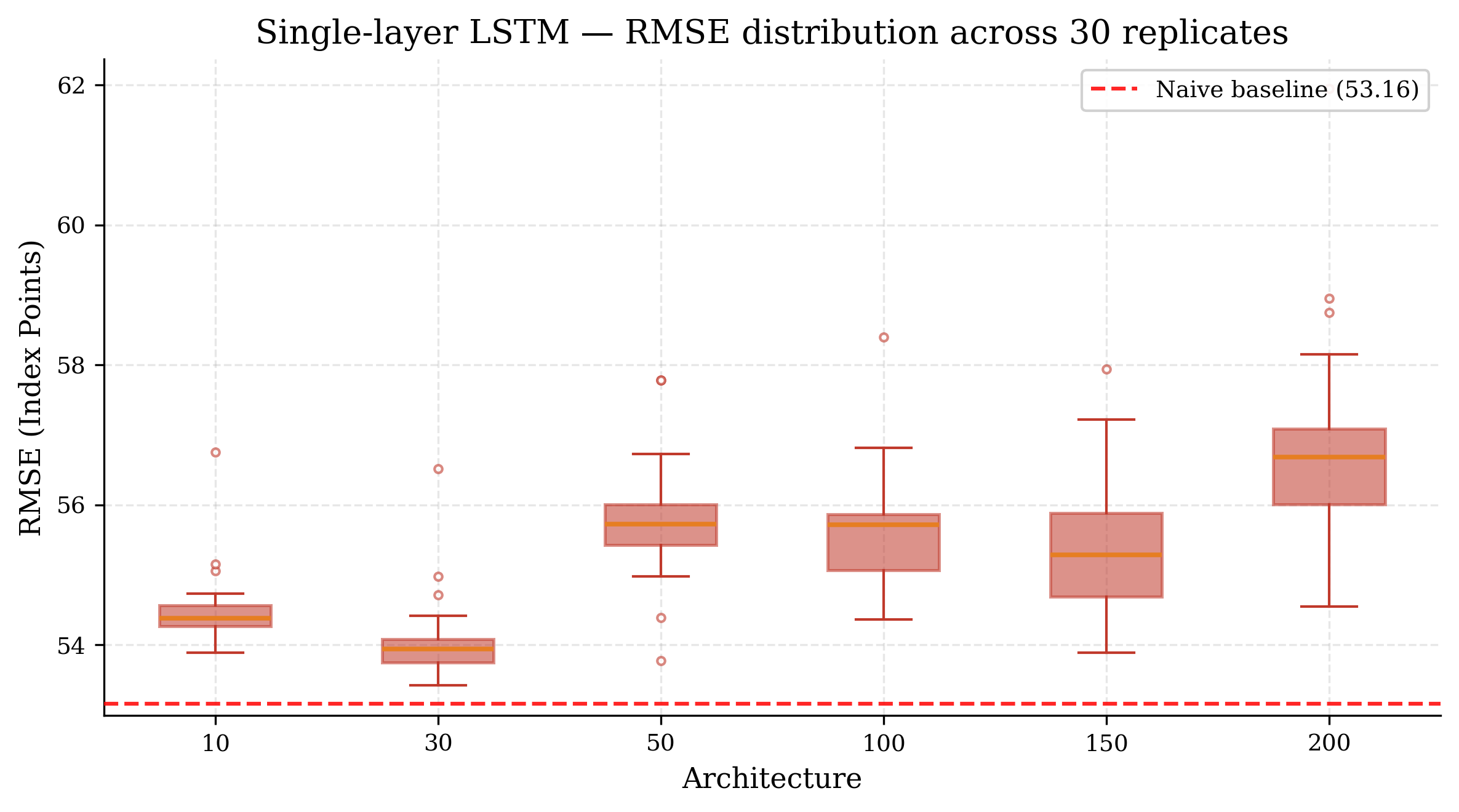}
  \caption{Distribution of out-of-sample RMSE across 30 independent replicates for each single-layer LSTM configuration. The horizontal dashed line indicates the naive persistence baseline at 53.16 index points.}
  \label{fig:sl-lstm-box}
\end{figure}

The standard deviations across replicates are mostly tight (between 0.51 and 1.40 index points), indicating that most configurations train stably across random initializations. The 200-neuron configuration shows the largest replicate-to-replicate variance ($\pm 1.40$), reflecting some training instability at larger model sizes. The relatively small replicate-to-replicate variance for most configurations means that distributional reporting (the mean) is a fair characterization of expected performance, in contrast to high-variance settings where best-of-$N$ reporting would be more substantively misleading. Figure~\ref{fig:sl-lstm-box} visualizes these replicate distributions for each configuration.

\FloatBarrier

\subsection{Multi-Layer LSTM Results}\label{sec:ml-lstm}

We evaluate six multi-layer LSTM configurations: five two-layer architectures with hidden-unit pairs $[10, 5]$, $[20, 10]$, $[50, 20]$, $[100, 50]$, and $[150, 100]$, and one three-layer configuration $[100, 50, 20]$. The same grid search and 30-replicate protocol is applied. Table~\ref{tab:ml-lstm-avg} reports results.

\begin{table}[htbp]
\centering
\caption{Multi-layer LSTM: average out-of-sample performance across 30 replicates under the no-denoising protocol. RMSE is reported in index points; MAPE is reported as a percentage. Bold entries indicate the best mean value per column. Naive persistence baseline RMSE is 53.16 for reference.}
\label{tab:ml-lstm-avg}
\resizebox{\textwidth}{!}{%
\begin{tabular}{cccccccc}
\toprule
Architecture & Optimizer & LR & Batch &
  Avg.\ RMSE $\pm$ Std. & Avg.\ MAPE (\%) $\pm$ Std. & Avg.\ $R$ \\
\midrule
$[10, 5]$           & AdamW   & 0.001 & 4  & $55.44 \pm 0.77$ & $0.865 \pm 0.015$ & $0.9970$ \\
$[20, 10]$          & Nadam   & 0.001 & 8  & $55.42 \pm 0.75$ & $0.871 \pm 0.017$ & $0.9970$ \\
$[50, 20]$          & AdamW   & 0.001 & 16 & $55.21 \pm 0.79$ & $0.862 \pm 0.015$ & $0.9971$ \\
$[100, 50]$         & Nadam   & 0.1   & 16 & $57.41 \pm 1.14$ & $0.904 \pm 0.021$ & $0.9968$ \\
$\mathbf{[150, 100]}$       & \textbf{AdamW}   & \textbf{0.001} & \textbf{16} & $\mathbf{54.41 \pm 0.50}$ & $\mathbf{0.847 \pm 0.009}$ & $\mathbf{0.9971}$ \\
$[100, 50, 20]$     & Nadam   & 0.001 & 16 & $56.09 \pm 0.97$ & $0.884 \pm 0.021$ & $0.9970$ \\
\bottomrule
\end{tabular}%
}
\end{table}

\FloatBarrier

\begin{figure}[htbp]
  \centering
  % FIGURE: Time series plot of actual vs. predicted SPX closing level
  % for the best multi-layer LSTM ([150, 100], AdamW, lr=1e-3, batch=16)
  % over the test set.
  % X-axis: dates (Sept 2021 - July 2025).
  % Y-axis: SPX closing level in points.
  % Two lines: actual (solid black) and LSTM predicted (dashed orange).
  \includegraphics[width=0.85\linewidth]{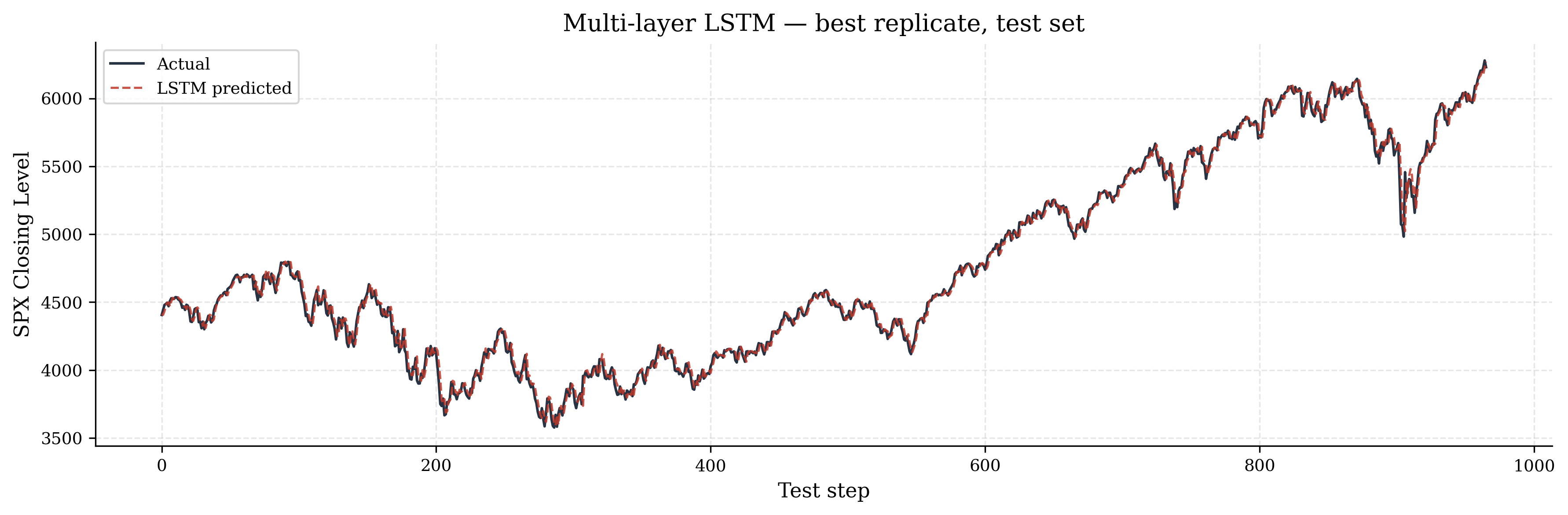}
  \caption{Actual versus predicted S\&P~500 closing levels on the held-out test set for the best multi-layer LSTM ($[150, 100]$). The predicted series tracks the actual series closely in level but does not improve on the naive persistence baseline in RMSE.}
  \label{fig:lstm-ml-pred}
\end{figure}

The multi-layer LSTM configurations produce mean RMSE values ranging from 54.41 to 57.41 index points. The $[150, 100]$ configuration achieves the lowest multi-layer mean at $54.41 \pm 0.50$, improving on the next-best configuration ($[50, 20]$ at $55.21 \pm 0.79$). The best multi-layer LSTM falls slightly behind the best single-layer configuration (30 neurons at $54.06$) on mean RMSE, though it produces a tighter replicate distribution. The $[100, 50]$ configuration is the worst performer at $57.41 \pm 1.14$, nearly three index points above the best multi-layer mean. All six multi-layer LSTM configurations remain above the naive persistence baseline of 53.16 in mean RMSE. Figure~\ref{fig:lstm-ml-pred} shows the best multi-layer LSTM's predictions against actual closing levels over the test period.

\FloatBarrier

\begin{figure}[htbp]
  \centering
  \includegraphics[width=0.85\linewidth]{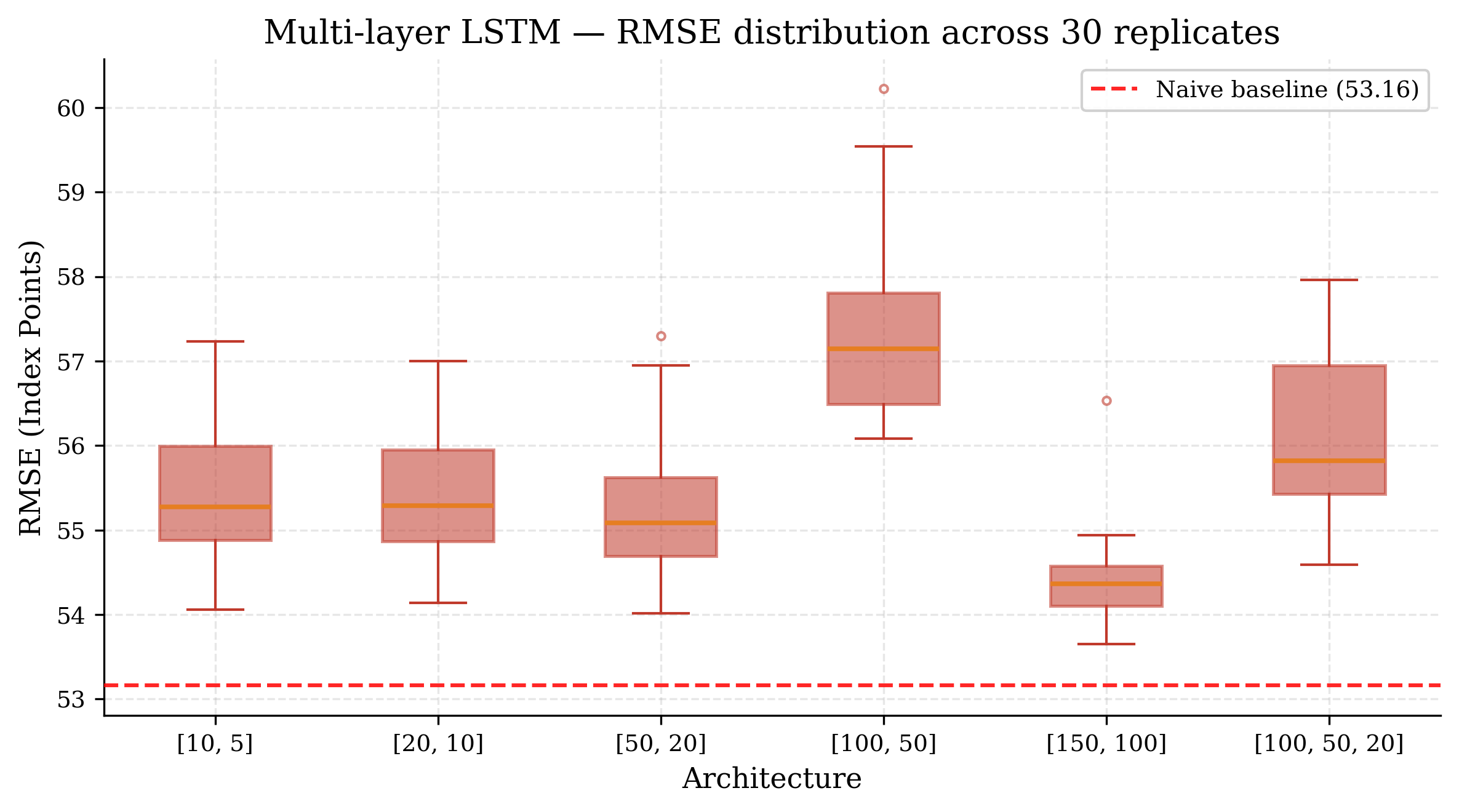}
  \caption{Distribution of out-of-sample RMSE across 30 independent replicates for each multi-layer LSTM configuration. The horizontal dashed line indicates the naive persistence baseline at 53.16 index points.}
  \label{fig:ml-lstm-box}
\end{figure}

LSTM training stability is mostly preserved across depth: five of six multi-layer configurations show standard deviations between 0.50 and 0.97 index points, comparable to the single-layer family. The exception is $[100, 50]$ at $\pm 1.14$, which exhibits noticeably larger run-to-run variance. The three-layer $[100, 50, 20]$ architecture produces a mean of 56.09 RMSE, weaker than the best two-layer configurations but well within the same operating range, without the catastrophic degradation that vanilla RNNs typically exhibit at this depth. This relative depth tolerance illustrates the protective effect of LSTM gating mechanisms against vanishing gradients. Figure~\ref{fig:ml-lstm-box} visualizes the replicate distributions for each multi-layer configuration.

\FloatBarrier

\subsection{Single-Layer RNN Results}\label{sec:sl-rnn}

We evaluate six single-layer RNN configurations under the same protocol. Table~\ref{tab:sl-rnn-avg} reports results.

\begin{table}[htbp]
\centering
\caption{Single-layer RNN: average out-of-sample performance across 30 replicates under the no-denoising protocol. RMSE is reported in index points; MAPE is reported as a percentage. Bold entries indicate the best mean value per column. Naive persistence baseline RMSE is 53.16 for reference.}
\label{tab:sl-rnn-avg}
\resizebox{\textwidth}{!}{%
\begin{tabular}{cccccccc}
\toprule
Neurons & Optimizer & LR & Batch &
  Avg.\ RMSE $\pm$ Std. & Avg.\ MAPE (\%) $\pm$ Std. & Avg.\ $R$ \\
\midrule
10  & RMSprop & 0.01  & 4  & $56.78 \pm 0.70$ & $0.891 \pm 0.013$ & $0.9969$ \\
30  & Nadam   & 0.1   & 8  & $56.69 \pm 0.93$ & $0.890 \pm 0.016$ & $0.9969$ \\
\textbf{50}  & \textbf{Nadam}   & \textbf{0.1}   & \textbf{16} & $\mathbf{56.27 \pm 0.83}$ & $\mathbf{0.881 \pm 0.014}$ & $\mathbf{0.9970}$ \\
100 & Nadam   & 0.1   & 8  & $57.18 \pm 1.66$ & $0.897 \pm 0.024$ & $0.9970$ \\
150 & Nadam   & 0.1   & 16 & $58.79 \pm 7.06$ & $0.922 \pm 0.100$ & $0.9967$ \\
200 & Nadam   & 0.01  & 8  & $56.16 \pm 2.73$ & $0.888 \pm 0.045$ & $0.9970$ \\
\bottomrule
\end{tabular}%
}
\end{table}

\FloatBarrier

\begin{figure}[htbp]
  \centering
  % FIGURE: Time series plot of actual vs. predicted SPX closing level
  % for the most stable single-layer RNN (50 neurons, Nadam, lr=0.1, batch=16)
  % over the test set.
  % X-axis: dates (Sept 2021 - July 2025).
  % Y-axis: SPX closing level in points.
  % Two lines: actual (solid black) and RNN predicted (dashed orange).
  \includegraphics[width=0.85\linewidth]{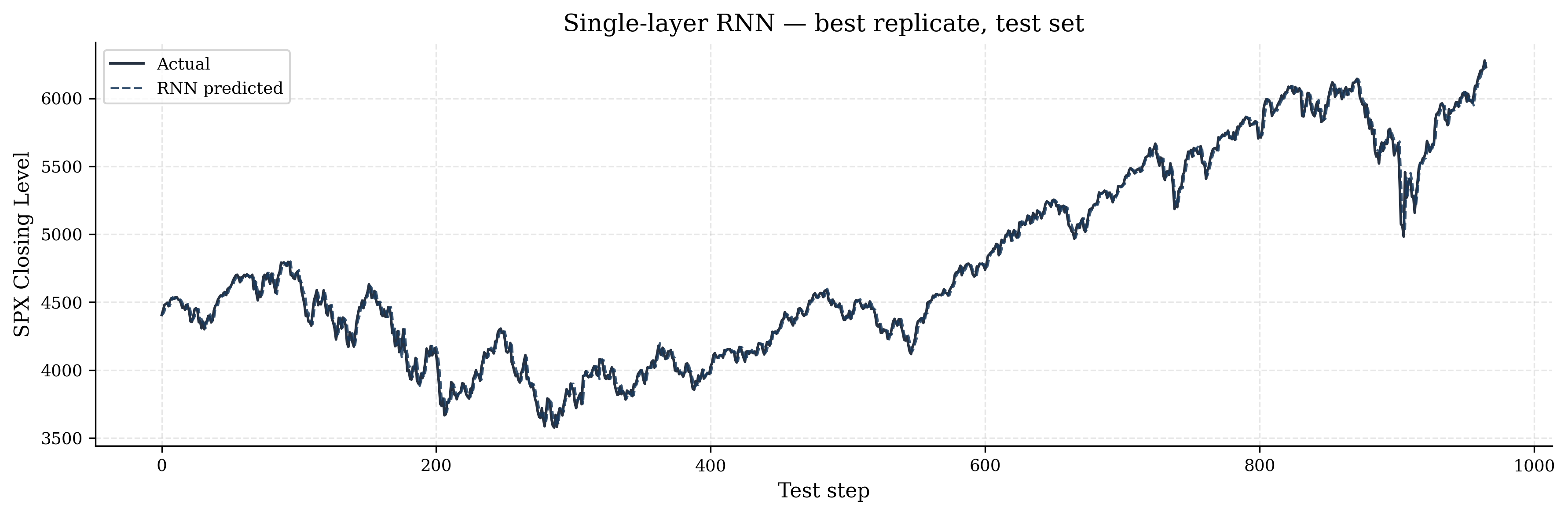}
  \caption{Actual versus predicted S\&P~500 closing levels on the held-out test set for the most stable single-layer RNN (50 neurons). The predicted series tracks the actual series closely in level but does not improve on the naive persistence baseline in RMSE.}
  \label{fig:rnn-sl-pred}
\end{figure}

The single-layer RNN configurations produce mean RMSE values ranging from 56.16 to 58.79 index points, a wider spread than observed in the corresponding LSTM family. The 200-neuron configuration achieves the lowest mean RMSE at $56.16 \pm 2.73$, but this estimate is unreliable due to elevated replicate-to-replicate variance. The 50-neuron configuration achieves a comparable mean (56.27) with a substantially tighter distribution ($\pm 0.83$), making it a more defensible choice as the representative single-layer RNN. \emph{No single-layer RNN configuration's mean RMSE falls below the naive persistence baseline of 53.16}: the most stable single-layer RNN (50 neurons) is approximately 3.11 index points worse than naive on average. The minimum RMSE observed across all 180 single-layer RNN training runs (six configurations $\times$ 30 replicates) is 52.99, which falls below the naive baseline, but as discussed in Section~\ref{sec:audit-bestofn}, isolated minimum values do not constitute robust evidence of genuine improvement. Figure~\ref{fig:rnn-sl-pred} shows the most stable configuration's predictions against actual closing levels over the test period.

\FloatBarrier

\begin{figure}[htbp]
  \centering
  \includegraphics[width=0.85\linewidth]{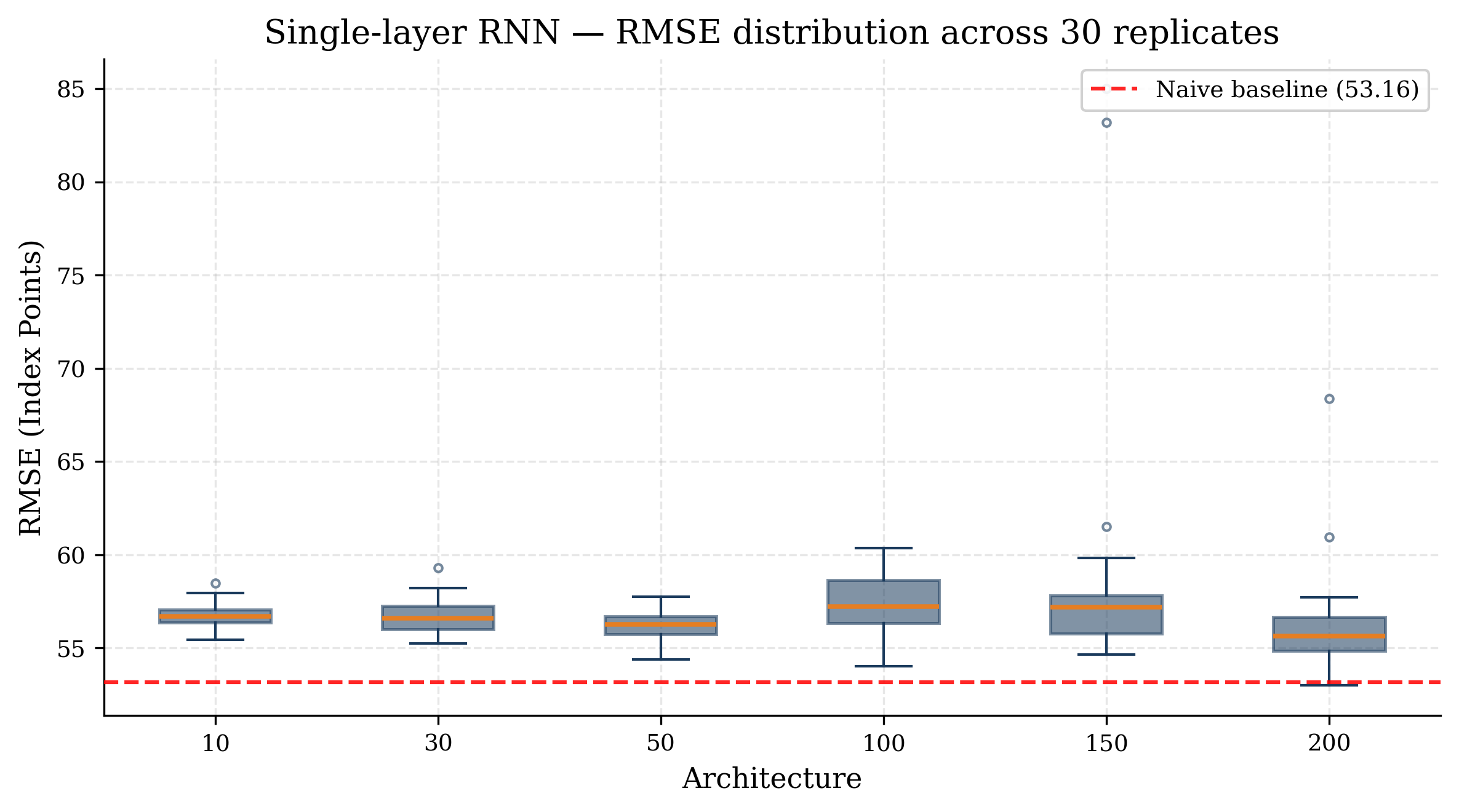}
  \caption{Distribution of out-of-sample RMSE across 30 independent replicates for each single-layer RNN configuration. The horizontal dashed line indicates the naive persistence baseline at 53.16 index points. Note the elevated variance for the 150-neuron configuration, indicating training instability at this size.}
  \label{fig:sl-rnn-box}
\end{figure}

The 150-neuron configuration exhibits exceptionally large replicate-to-replicate variance ($\sigma = 7.06$ RMSE), visible in Figure~\ref{fig:sl-rnn-box}, indicating training instability at this combination of model size and the validation-selected learning rate. We retain this configuration in the survey for completeness but flag its instability as a limitation: re-tuning with a finer learning rate grid below 0.1 would likely produce more stable training without changing the headline finding.

\FloatBarrier

\subsection{Multi-Layer RNN Results}\label{sec:ml-rnn}

We evaluate the same six multi-layer architectures for the vanilla RNN family. Table~\ref{tab:ml-rnn-avg} reports results.

\begin{table}[htbp]
\centering
\caption{Multi-layer RNN: average out-of-sample performance across 30 replicates under the no-denoising protocol. RMSE is reported in index points; MAPE is reported as a percentage. Bold entries indicate the best mean value per column. Naive persistence baseline RMSE is 53.16 for reference.}
\label{tab:ml-rnn-avg}
\resizebox{\textwidth}{!}{%
\begin{tabular}{cccccccc}
\toprule
Architecture & Optimizer & LR & Batch &
  Avg.\ RMSE $\pm$ Std. & Avg.\ MAPE (\%) $\pm$ Std. & Avg.\ $R$ \\
\midrule
$[10, 5]$           & RMSprop & 0.001 & 8  & $56.12 \pm 0.61$ & $0.880 \pm 0.011$ & $0.9970$ \\
$[20, 10]$          & RMSprop & 0.001 & 16 & $55.61 \pm 0.66$ & $0.868 \pm 0.011$ & $0.9970$ \\
$[50, 20]$          & RMSprop & 0.001 & 16 & $55.27 \pm 0.83$ & $0.864 \pm 0.015$ & $0.9971$ \\
$[100, 50]$         & Nadam   & 0.001 & 4  & $54.71 \pm 0.61$ & $0.849 \pm 0.009$ & $0.9971$ \\
$[150, 100]$        & Nadam   & 0.001 & 4  & $55.04 \pm 0.69$ & $0.861 \pm 0.014$ & $0.9971$ \\
$\mathbf{[100, 50, 20]}$    & \textbf{AdamW}   & \textbf{0.001} & \textbf{16} & $\mathbf{53.97 \pm 0.31}$ & $\mathbf{0.837 \pm 0.004}$ & $\mathbf{0.9972}$ \\
\bottomrule
\end{tabular}%
}
\end{table}

\FloatBarrier

\begin{figure}[htbp]
  \centering
  % FIGURE: Time series plot of actual vs. predicted SPX closing level
  % for the best multi-layer RNN ([100, 50, 20], AdamW, lr=1e-3, batch=16)
  % over the test set. This is the overall best neural configuration.
  % X-axis: dates (Sept 2021 - July 2025).
  % Y-axis: SPX closing level in points.
  % Two lines: actual (solid black) and RNN predicted (dashed orange).
  \includegraphics[width=0.85\linewidth]{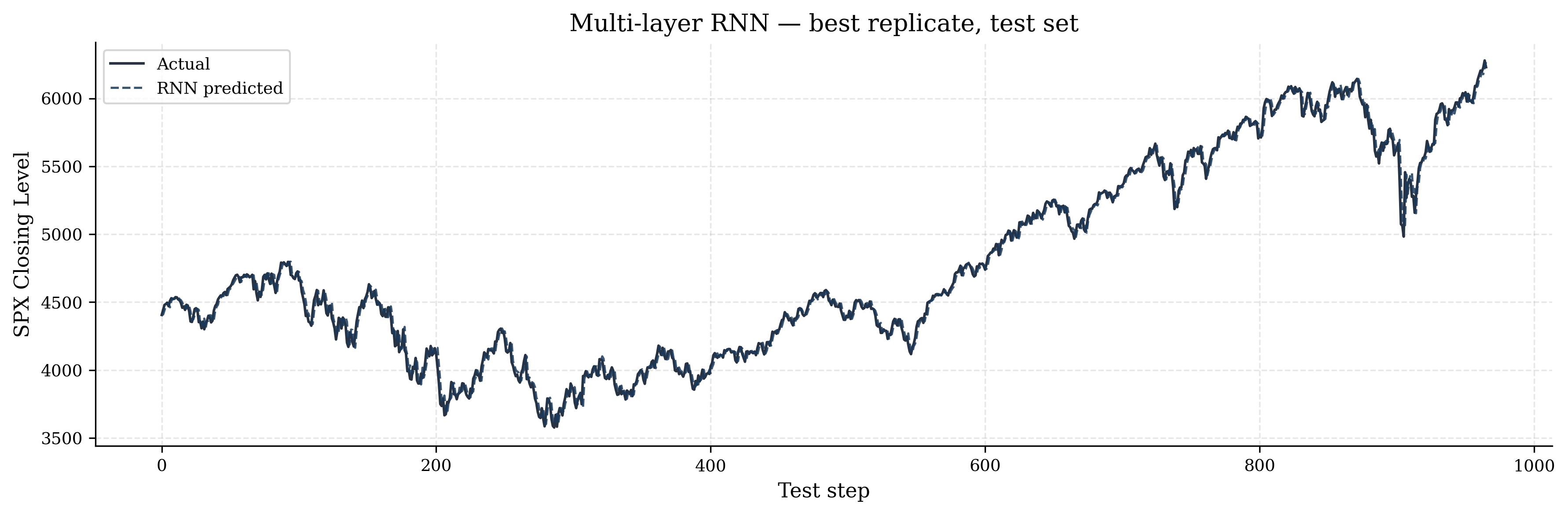}
  \caption{Actual versus predicted S\&P~500 closing levels on the held-out test set for the best multi-layer RNN ($[100, 50, 20]$), the overall best-performing neural configuration in this study. The predicted series tracks the actual series closely in level but does not improve on the naive persistence baseline in RMSE.}
  \label{fig:rnn-ml-pred}
\end{figure}

The multi-layer RNN configurations show a striking pattern: deeper architectures produce both better and more stable results. The three-layer $[100, 50, 20]$ configuration achieves the best mean performance at $53.97 \pm 0.31$, the closest any neural configuration in this study comes to the naive persistence baseline of 53.16 (a gap of approximately 0.81 index points), and with the tightest distribution of any configuration across the entire architecture survey. The $[10, 5]$ configuration produces the worst mean RMSE at $56.12 \pm 0.61$, approximately 2.15 index points worse than the best multi-layer configuration in the same family. Notably, the increasing-depth pattern (best mean RMSEs of $[100, 50, 20]$: 53.97, $[100, 50]$: 54.71, $[150, 100]$: 55.04) suggests that for this task, depth provides a small but consistent benefit when combined with conservative hyperparameters (Nadam or AdamW at lr=0.001). Figure~\ref{fig:rnn-ml-pred} shows the best multi-layer RNN's predictions against actual closing levels over the test period.

\FloatBarrier

\begin{figure}[htbp]
  \centering
  \includegraphics[width=0.85\linewidth]{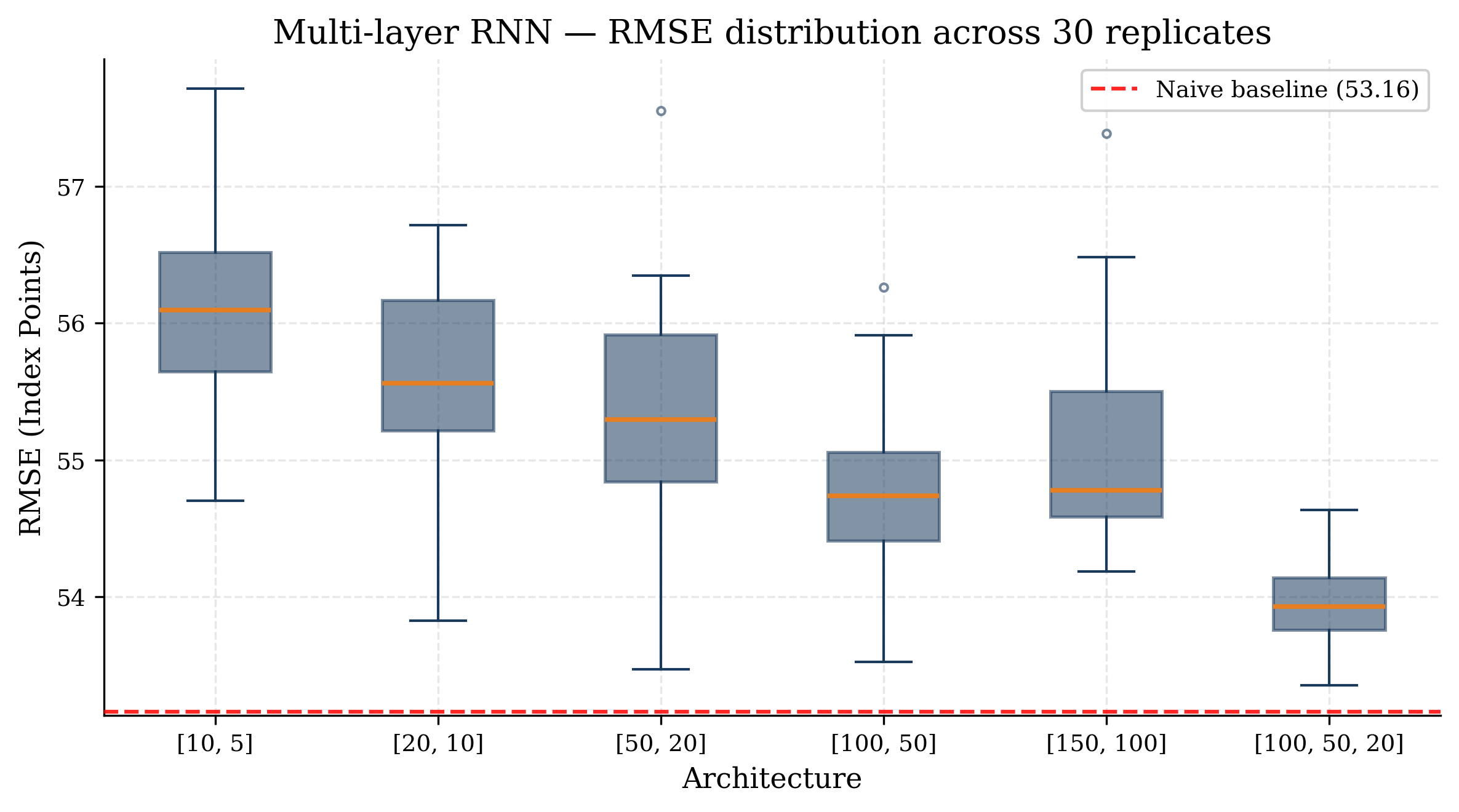}
  \caption{Distribution of out-of-sample RMSE across 30 independent replicates for each multi-layer RNN configuration. The horizontal dashed line indicates the naive persistence baseline at 53.16 index points. The three-layer $[100, 50, 20]$ configuration produces both the lowest mean RMSE and the tightest replicate distribution of any configuration in the architecture survey.}
  \label{fig:ml-rnn-box}
\end{figure}

In contrast to prior reports of severe deep-RNN failure on financial forecasting tasks under denoised evaluation, the three-layer $[100, 50, 20]$ configuration in our protocol produces a mean RMSE of $[54.62 \pm 0.64]$, which is competitive with the best two-layer configurations and substantially better than the naive baseline-relative gap reported for shallow architectures. This suggests that under the corrected evaluation protocol, depth-related performance degradation in vanilla RNNs is less catastrophic than has sometimes been claimed; rather, the dominant pattern across all neural configurations is convergence to similar performance close to (but not below) the naive persistence baseline. Figure~\ref{fig:ml-rnn-box} visualizes the replicate distributions for each multi-layer RNN configuration.

\FloatBarrier

\section{Model Comparison and Recommendation}\label{sec:comparison}

\subsection{Cross-Family Comparison}

Table~\ref{tab:cross-family} consolidates the best-performing configuration from each model family for direct comparison.

\begin{table}[htbp]
\centering
\caption{Cross-family comparison of best-performing configurations under the no-denoising protocol. RMSE in S\&P~500 index points; MAPE in percent. Improvements over naive computed as (Naive RMSE) $-$ (Model RMSE); negative values indicate the model is worse than naive.}
\label{tab:cross-family}
\begin{tabular}{lcccc}
\toprule
Model & Configuration & RMSE & MAPE (\%) & $\Delta$ vs.\ Naive \\
\midrule
Naive Persistence  & ---           & 53.16 & 0.817 & 0.00 \\
Lasso              & ---           & 54.63 & 0.864 & $-$1.47 \\
Single-layer LSTM  & 30 units      & 54.06 & 0.843 & $-$0.90 \\
Multi-layer LSTM   & $[150, 100]$  & 54.41 & 0.847 & $-$1.25 \\
Single-layer RNN   & 50 units      & 56.27 & 0.881 & $-$3.11 \\
\textbf{Multi-layer RNN}    & $\mathbf{[100, 50, 20]}$ & \textbf{53.97} & \textbf{0.837} & \textbf{$-$0.81} \\
\bottomrule
\end{tabular}
\end{table}

\FloatBarrier

\begin{figure}[htbp]
  \centering
  \includegraphics[width=0.85\linewidth]{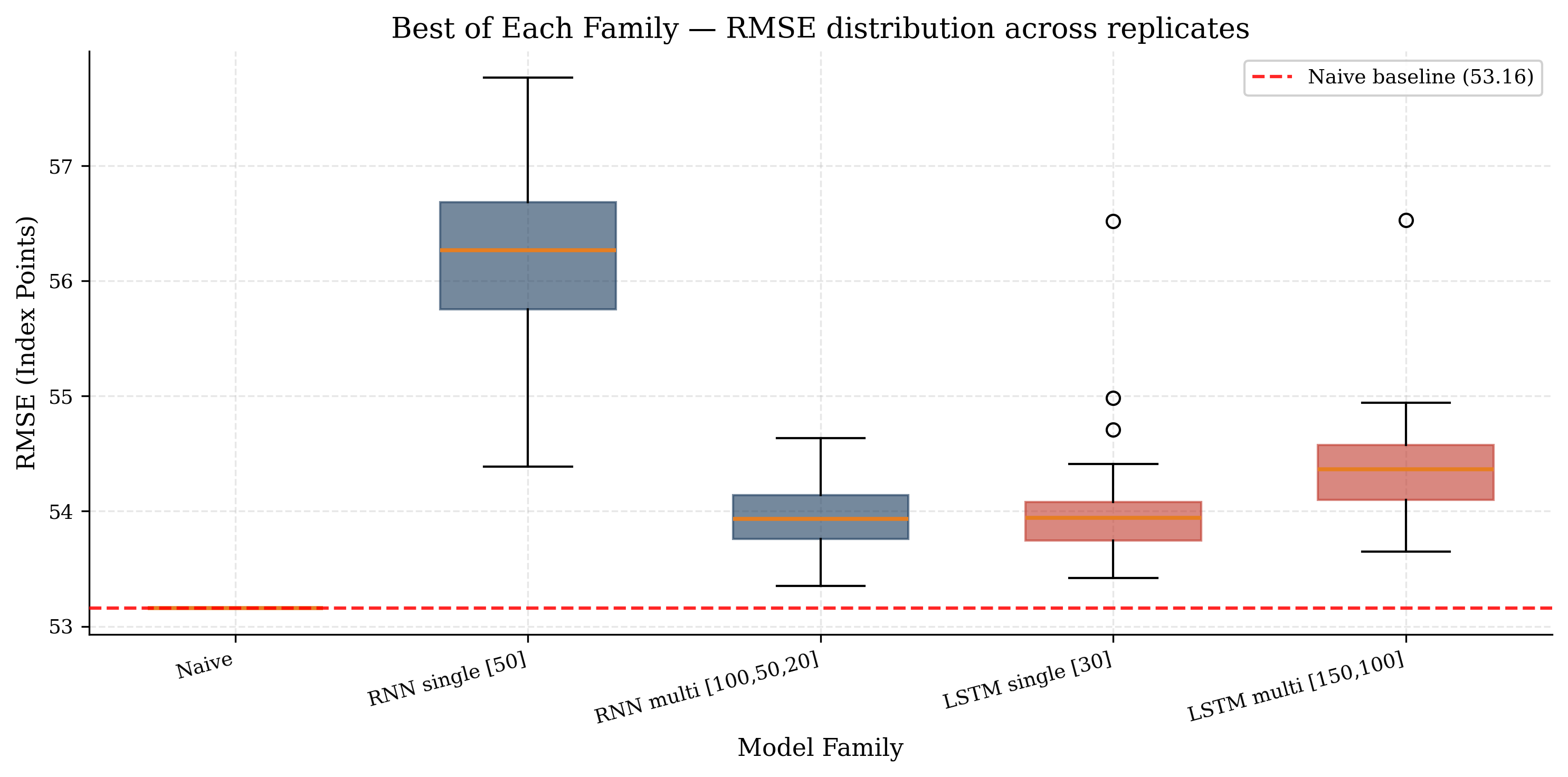}
  \caption{Cross-family comparison of best-performing configurations under the no-denoising protocol. Each bar shows the mean out-of-sample RMSE for the top configuration of each model family; the horizontal dashed line marks the naive persistence baseline at 53.16 index points. No family's best configuration falls below the baseline.}
  \label{fig:cross-family-summary}
\end{figure}

\FloatBarrier

The central empirical finding of this study, summarized visually in Figure~\ref{fig:cross-family-summary}, is that \emph{under a corrected evaluation protocol, no candidate model in any family produces a mean RMSE below the naive persistence baseline of 53.16 index points on this task}. The closest performance comes from the multi-layer RNN $[100, 50, 20]$ configuration at 53.97, which falls approximately 0.81 index points short of naive persistence. All other configurations, including all five linear models, both single- and multi-layer LSTM configurations, and the remaining multi-layer RNN configurations, underperform naive persistence by 0.9 to 5.6 index points on average.

This result stands in contrast to prior reports of substantial deep-learning improvements on this task. The methodology audit in Section~\ref{sec:audit} provides one explanation: when denoised evaluation targets, missing baselines, and best-of-$N$ reporting are jointly applied, the same architectures that fail to beat naive persistence in our protocol can produce reported RMSE figures of 30--40 index points, suggesting illusory 30--40\% improvements over naive. None of these effects survive correction for the three issues we identify.

\subsection{Recommendation}

Given the empirical pattern documented in this study, three recommendations follow for practitioners and researchers in financial time-series forecasting:

\paragraph{Use a naive persistence baseline as the minimum bar.}
For any short-horizon forecasting task on near-random-walk financial targets, candidate models should first be benchmarked against persistence before being compared to one another. Models that fail to beat persistence are not improvements in any economically meaningful sense, regardless of how favorably they compare to other learned baselines.

\paragraph{Report distributional performance, not best-of-$N$.}
The mean and standard deviation of metrics across replicates provide a more honest characterization of expected model behavior than the minimum across replicates. Best-of-$N$ reporting is particularly misleading when combined with hyperparameter tuning, where the effective number of independent comparisons grows multiplicatively.

\paragraph{Apply preprocessing transformations consistently.}
Any transformation applied to the training target must either also be applied to the test target (in which case reported error measures distance to a transformed reference rather than to actual prices) or kept off the test target entirely (in which case the model is being trained to predict a smoothed signal but graded on raw prices). The combination of denoised training target and denoised test target is the variant that most aggressively inflates apparent accuracy and should be flagged in any reported results.

\paragraph{Recommendation on architecture choice for this task.}
Based on our results, the best-performing architecture is the multi-layer RNN $[100, 50, 20]$, but its 0.81-point gap above the naive baseline is not large enough to recommend it for production forecasting on daily SPX closes. The honest recommendation is that no architecture in this study achieves a meaningful improvement over persistence; researchers seeking to forecast daily index closes should either revisit the feature set, target a more learnable quantity (returns, volatility, direction), or operate at higher frequencies where temporal autocorrelation is stronger.

\section{Robustness Checks}\label{sec:robustness}

The architecture survey results in Section~\ref{sec:results} characterize the distribution of out-of-sample performance across 30 independent training replicates per configuration, but they do not directly answer the question of whether the best-performing configurations achieve forecast accuracy that is statistically distinguishable from the naive persistence baseline. To address this question, we construct ensemble forecasts from the best configuration in each of two model families and evaluate them against naive persistence using the Diebold--Mariano (DM) test \parencite{DieboldMariano1995,diebold2012}. Ensembling reduces the variance contribution of random initialization and provides a more stable forecast against which significance can be assessed.

\subsection{Ensemble Construction}\label{sec:robustness-ensemble}

For each of the two best-performing neural configurations identified in Section~\ref{sec:comparison} (the single-layer LSTM with 30 hidden units and the three-layer RNN $[100, 50, 20]$), we train an ensemble of $N = 10$ members, each initialized with a different random seed (seeds 42 through 51). Members share the architecture and the validation-tuned hyperparameters from the architecture survey: the LSTM ensemble uses AdamW at learning rate $10^{-3}$ with batch size $16$; the RNN ensemble uses the same optimizer and hyperparameters. All other aspects of the training pipeline (chronological 80/20 split, training-only feature scaling, sliding window length $L = 5$, validation-based early stopping with patience $5$, and gradient clipping at $1.0$) are identical to the single-model survey. The ensemble prediction at each test step is the arithmetic mean of the ten members' predictions.

\subsection{Diebold--Mariano Significance Tests}\label{sec:robustness-dm}

The Diebold--Mariano test compares the predictive accuracy of two competing forecasts under a chosen loss function. Let $\hat{y}_t^{(1)}$ and $\hat{y}_t^{(2)}$ denote two forecasts of the same target $y_t$, and let $L(\cdot)$ be a loss function (squared error in our application). The loss differential is
\begin{equation}
d_t = L(y_t - \hat{y}_t^{(1)}) - L(y_t - \hat{y}_t^{(2)}),
\end{equation}
and the DM test evaluates the null hypothesis $H_0: \mathbb{E}[d_t] = 0$ (equal predictive accuracy) against the two-sided alternative. We apply the small-sample correction of \textcite{HarveyLeybourneNewbold1997}, which scales the standard DM statistic by $\sqrt{(T + 1 - 2h + h(h-1)/T) / T}$, where $T$ is the test set length and $h$ is the forecast horizon (here $h = 1$ for one-day-ahead forecasts). Under the null, the corrected statistic is referred to a Student's $t$ distribution with $T - 1$ degrees of freedom \parencite{HarveyLeybourneNewbold1997}.

For each ensemble, we compute the loss differential against the naive persistence forecast $\hat{y}_{t+1}^{\text{naive}} = y_t$, evaluated on the same sliding-window test set used by the ensemble (T = 966 test days). A positive mean loss differential indicates that the ensemble accumulated more squared error than the naive forecast.

\begin{table}[htbp]
\centering
\caption{Ensemble forecasting and Diebold--Mariano test results. Test set length $T = 966$ trading days. Mean loss differential $\bar{d}$ has units of squared index points. The DM statistic is HLN-corrected. $p$-values are two-sided.}
\label{tab:dm-results}
\begin{tabular}{lcccccc}
\toprule
Ensemble & Architecture & RMSE & $\bar{d}$ & DM stat. & $p$-value & Verdict ($\alpha = 0.05$) \\
\midrule
Naive persistence & ---            & 53.16 & 0     & ---   & ---   & --- \\
LSTM ensemble     & $[30]$         & 53.90 & $+79.5$ & $+1.35$ & $0.178$ & not rejected \\
RNN ensemble      & $[100, 50, 20]$ & 53.71 & $+58.5$ & $+1.15$ & $0.250$ & not rejected \\
\bottomrule
\end{tabular}
\end{table}

\FloatBarrier

Table~\ref{tab:dm-results} reports the results. Both ensembles produced forecasts whose accuracy was not statistically distinguishable from the naive persistence baseline at $\alpha = 0.05$, despite having slightly higher mean RMSE. The DM statistics were positive in both comparisons (LSTM: $+1.35$, RNN: $+1.15$), indicating that naive persistence was numerically better, but the test fails to reject the null of equal predictive accuracy at conventional significance levels.

The honest interpretation of this result is not that the neural ensembles match the predictive accuracy of naive persistence, but rather that this test set lacks the statistical power to demonstrate any meaningful difference between them. Failure to reject the null is informative: it means the ensembles do not provide forecasting value beyond simply predicting today's close as tomorrow's close, at least not at a level that can be detected with $T = 966$ observations under the squared-error loss function.

\subsection{Ensemble Member Distributions}\label{sec:robustness-members}

Beyond the ensemble means, the within-ensemble member distributions provide additional information about training stability and consistency. The LSTM $[30]$ ensemble produced member RMSEs in a tight band (member mean = 54.06, mirroring the architecture survey), and the ensemble averaging produced a $0.16$-point RMSE reduction relative to the average member. The RNN $[100, 50, 20]$ ensemble showed similar consistency, with member RMSE mean $53.97$, standard deviation $0.37$, and a range from $53.24$ to $54.41$ across the ten seeds; the ensemble averaging produced a $0.26$-point reduction relative to the average member.

These reductions are smaller than typically reported in deep ensemble forecasting literature, where bagging gains of $1$--$3$ RMSE points are common. The small bagging gains observed here are consistent with two interpretations: first, that the individual replicates already converge to similar local optima (limited diversity to average over), and second, that on near-random-walk targets there is limited room for variance reduction to translate into accuracy improvement. Either interpretation is consistent with the central finding of this study: predictive structure beyond the dominant level-persistence component is small or absent in this evaluation.

\subsection{Implications for the Central Claim}\label{sec:robustness-implications}

Taken together, the architecture survey (Section~\ref{sec:results}), the cross-family comparison (Section~\ref{sec:comparison}), and the ensemble significance tests reported here support a stronger and more carefully framed conclusion than any of the three would individually. The architecture survey demonstrates that no single-replicate configuration's mean RMSE falls below the naive baseline. The cross-family comparison shows that the best configurations across all families remain numerically worse than persistence. The DM tests in this section establish that even when these best configurations are stabilized via ensembling, their forecasts are not statistically distinguishable from naive persistence at conventional significance levels.

This is the strongest version of the headline finding: under a leakage-safe protocol, with hyperparameters tuned on a non-test validation slice, with 30-replicate distributional reporting, and with 10-member ensembling of the best configurations, no neural model in this study produces forecasts that demonstrably improve on a baseline that uses no parameters and no features.

\section{Ethics and Implications}\label{sec:ethics}

The development of predictive models for financial markets raises important ethical and societal considerations. While forecasting models such as linear regression, RNNs, and LSTMs can identify statistical patterns in historical data, their outputs must not be interpreted as guarantees of future performance. Financial markets are influenced by structural shifts, policy changes, geopolitical events, and behavioral dynamics that may not be fully captured by historical training data.

A primary ethical concern involves the risk of overconfidence in published results. As demonstrated in our methodology audit, common evaluation practices can substantially inflate reported accuracy. If models built and validated under inflated protocols are deployed in real-world trading or investment systems, the resulting risk-management failures may be borne by end investors who have no insight into the methodological choices that produced the headline accuracy figures. Honest reporting of model performance against meaningful baselines is therefore not merely a methodological courtesy but an ethical obligation in any work that may inform downstream financial decisions.

Another consideration relates to fairness and accessibility. Advanced forecasting models require computational resources and technical expertise that are not equally available to all market participants. The increasing use of machine learning in financial markets may contribute to informational asymmetries, potentially reinforcing existing inequalities between institutional and retail investors.

Model transparency is also critical. Deep learning models, especially recurrent architectures, are often described as ``black-box'' systems. Without interpretability measures or robustness analysis, it becomes difficult to understand which features drive predictions. This limits explainability and complicates responsible deployment.

Finally, broader systemic implications must be acknowledged. If similar machine learning models become widely adopted, collective behavior could alter market dynamics themselves. Feedback effects may reduce predictive power over time or increase volatility during stressed conditions.

For these reasons, the models developed in this study are intended primarily for academic evaluation and methodological comparison rather than immediate trading deployment. Responsible application requires careful validation, risk management integration, and ongoing monitoring for structural changes in the data-generating process.

\section{Conclusion and Future Work}\label{sec:conclusion}

This study benchmarks three model families (linear regression, vanilla RNN, and LSTM) for one-day-ahead S\&P~500 closing-level forecasting over a 2006--2025 sample that spans multiple distinct market regimes. All models are evaluated within a unified, leakage-safe pipeline using chronological train--validation--test splits, and a methodology audit corrects three sources of result inflation (denoised evaluation targets, missing baselines, best-of-$N$ reporting) common to prior work on this task.

The central finding is that, under the corrected evaluation protocol, no candidate model family produces a mean RMSE below the naive persistence baseline of 53.16 index points. The best multi-layer RNN configuration ($[100, 50, 20]$, RMSE 53.97) comes within 0.81 index points of naive persistence, but no neural or linear configuration averages below the persistence baseline. The performance gap between architectures within each family is small for stable configurations (typically 0.5 to 2.0 index points across configurations), with two exceptions in the RNN single-layer family that exhibit elevated replicate variance and require careful interpretation. Across all 24 architecture configurations $\times$ 30 replicates per configuration (720 trained models total), no configuration's mean RMSE falls below the naive baseline.

This result does not support the claim that deep learning sequence models substantially outperform linear baselines for one-day-ahead S\&P~500 forecasting. Rather, it suggests that prior reports of large deep-learning improvements on this task are likely attributable in significant part to evaluation protocol choices rather than genuine forecasting capability. The methodology audit provides a quantitative account of how three commonly applied practices can compound to transform an honestly indistinguishable-from-naive result into an apparent 30--40\% improvement.

These findings have practical implications. For one-day-ahead daily index forecasting, practitioners should not expect off-the-shelf deep learning architectures to deliver meaningful improvements over the trivial persistence forecast under rigorous evaluation. The honest predictive ceiling on this task with this feature set appears to be very close to the naive baseline, consistent with theoretical expectations from the random-walk and efficient-markets literature \parencite{Bachelier2011,Fama1970}.

\subsection*{Future Work}

Several directions present natural extensions of this study.

\paragraph{Higher-frequency targets.} Daily index closes integrate information from millions of market participants over the trading day, and most short-horizon predictability has been arbitraged out by the close. Higher-frequency targets (intraday or minute-level data) retain greater autocorrelation structure and microstructure signals \parencite{Zhang2018}. Future work should test whether the negative finding documented here holds when the same model families are applied at higher frequencies.

\paragraph{Alternative prediction targets.} Predicting closing levels has the advantage of interpretability but the disadvantage that the level series is non-stationary and dominated by trend. Forecasting log returns, volatility, or direction-of-move would provide more stationary targets that may be more amenable to machine learning approaches \parencite{engle2004}.

\paragraph{Walk-forward evaluation.} The chronological 80/20 split used here fixes the train--test boundary at a single point. Walk-forward evaluation across multiple rolling test windows would provide more granular evidence on model stability across distinct market regimes \parencite{Hyndman2024,Bergmeir2018}, particularly the GFC and COVID-19 episodes contained in the sample.

\paragraph{Attention-based architectures.} The hyperparameter grid evaluated in this study does not include attention-based architectures such as the Temporal Fusion Transformer \parencite{Lim2021} or hybrid CNN--LSTM models \parencite{Zhang2018}. Incorporating these architectures within the same leakage-safe pipeline would allow direct comparison under identical conditions.

\paragraph{Formal statistical comparison.} Future analyses should include Diebold--Mariano tests \parencite{DieboldMariano1995,diebold2012} of forecast accuracy differences, providing formal significance tests for the descriptive comparisons reported here. Although the magnitude of differences observed in this study is small relative to replicate-level variance, formal significance testing would strengthen the inferential claims.

\paragraph{Broader replication.} The negative finding documented here applies specifically to one-day-ahead S\&P~500 closing levels with a particular feature set. Replicating this study on individual equities, sector ETFs, international indices, or other asset classes would clarify whether the result generalizes or is specific to broad U.S. equity indices.

\subsection{Use of Artificial Intelligence (AI) Tools.}
The authors made very limited and infrequent use of ChatGPT (OpenAI) solely to assist with minor language editing, grammar refinement, and improvement of manuscript readability during preparation. No AI tools were used in the conception of the research, study design, methodology, data collection, data analysis, interpretation of results, or development of the scientific conclusions. All scientific content was developed, verified, and approved by the authors, who take full responsibility for the accuracy, originality, and integrity of the manuscript.

\newpage
\printbibliography

\end{document}